%% file: acl_latex.tex
\pdfoutput=1

\documentclass[11pt]{article}

\usepackage[final]{acl}

\usepackage{times}
\usepackage{latexsym}
\usepackage{multirow}
\usepackage{booktabs}
\usepackage{multirow}
\usepackage{adjustbox}
\usepackage{subcaption} 
\usepackage{amsmath}
\usepackage{longtable}
\usepackage{booktabs}
\usepackage{ulem}
\usepackage[utf8]{inputenc}
\usepackage{newunicodechar}
\newunicodechar{₂}{$_2$}
\usepackage[T1]{fontenc}

\usepackage[utf8]{inputenc}

\usepackage{microtype}

\usepackage{inconsolata}

\usepackage{graphicx}

\title{Explaining Generalization of AI-Generated Text Detectors Through Linguistic Analysis}

\author{
 \textbf{Yuxi Xia\textsuperscript{1,2}},
 \textbf{Kinga Stańczak\textsuperscript{3}},
 \textbf{Benjamin Roth\textsuperscript{1,4}}
\\
 \textsuperscript{1}Faculty of Computer Science, University of Vienna, Vienna, Austria \\
 \textsuperscript{2}UniVie Doctoral School Computer Science, Vienna, Austria\\
  \textsuperscript{3}Department of Language Science and Technology, Saarland University, Germany\\
\textsuperscript{4}Faculty of Philological and Cultural Studies, University of Vienna, Vienna, Austria
\\
 \small{
   \textbf{Correspondence:} \href{mailto:email@domain}{yuxi.xia@univie.ac.at}
 }
}

\begin{document}
\maketitle
\begin{abstract}

AI-text detectors achieve high accuracy on in-domain benchmarks, but often struggle to generalize across different generation conditions such as unseen prompts, model families, or domains. While prior work has reported these generalization gaps, there are limited insights about the underlying causes. In this work, we present a systematic study aimed at explaining generalization behavior through linguistic analysis. We construct a comprehensive benchmark that spans 6 prompting strategies, 7 large language models (LLMs), and 4 domain datasets, resulting in a diverse set of human- and AI-generated texts. Using this dataset, we fine-tune classification-based detectors on various generation settings and evaluate their cross-prompt, cross-model, and cross-dataset generalization.
To explain the performance variance, we compute correlations between generalization accuracies and feature shifts of 80 linguistic features between training and test conditions. 
Our analysis reveals that generalization performance for specific detectors and evaluation conditions is significantly associated with linguistic features such as tense usage and pronoun frequency. \footnote{Code and data is available at: \url{https://github.com/Yuuxii/Generalization-of-AI-text-Detector}}

\end{abstract}
\section{Introduction}
The ability to reliably detect AI-generated text is becoming increasingly critical as large language models (LLMs) are deployed in education, media, and content moderation \cite{guo2024detective, hu2023radar}. While recent detectors achieve near-perfect performance on standard benchmarks \cite{guo2023close, wang-etal-2024-m4gt}, these evaluations typically assume that training and testing data are drawn from the same distribution. In real-world scenarios, however, AI-generated text varies widely across prompts, model families, and domains, raising serious concerns about how well detectors generalize under distribution shifts.

Prior studies have begun to examine generalization in the context of unseen prompts, models, or datasets \cite{xu-etal-2024-generalization, 10.1145/3658644.3670392}. However, these efforts largely focus on reporting performance drops without probing the underlying causes. Meanwhile, recent benchmark datasets introduce diversity in generation settings \cite{wang-etal-2024-m4gt, macko_multitude_2023, li_mage_2024}, but offer limited interpretability regarding the features detectors rely on. A more systematic and interpretable approach is needed to explain \textbf{why} generalization succeeds or fails.

In this paper, we propose to understand generalization through the lens of linguistic analysis. We hypothesize that changes in surface-level linguistic features, such as verb tense, syntactic complexity, or pronoun usage, can partially account for generalization behavior. To test this hypothesis, we construct a comprehensive benchmark combining 7 LLMs (e.g., Deepseek \cite{deepseekai2025deepseekr1incentivizingreasoningcapability}, Mistral \cite{mistral2024largeinstruct}), 4 domains (abstracts, news, reviews, QA), and 6 prompting strategies (e.g., few-shot, chain-of-thought (CoT)), enabling evaluation across prompt, model, and dataset generalization. We train the AI-text detectors by fine-tuning on two state-of-the-art models (XLM-RoBERTa \cite{conneau-etal-2020-unsupervised} and DeBERTa-V3 \cite{he2021debertav3})  for binary classification tasks. Each detector is trained with the texts generated by every possible condition (combination of prompt, model and dataset). The fine-tuned detectors are evaluated for cross-prompt, cross-model and cross-dataset generalization performance. Our results reveal substantial performance degradation under out-of-domain conditions, despite near-perfect in-domain accuracy.

To explain these generalization gaps, we perform a large-scale correlation analysis between detector generalization performance and score changes in 80 linguistic feature metrics across training and testing settings. We find that linguistic features have significant (p<0.05) correlations with generalization across different test settings.  While some linguistic features (e.g., passive voice, short sentence ratio) are strongly correlated (Pearson correlation > 0.7) with generalization behavior in specific configurations, there is no universal linguistic signal that explains all cases. Our findings suggest that linguistic features offer a useful, though partial, explanation of generalization, and that detectors may rely on different features depending on their training setup and the test conditions.

In summary, our main contributions are:  
(1) A new benchmark for evaluating AI-text detector generalization across 6 prompts, 4 domains, and 7 LLMs;  
(2) A comprehensive analysis linking linguistic feature shifts to detector generalization performance;  
(3) Insights into which linguistic features are most predictive of generalization behavior, helping to guide the development of more robust and interpretable detectors.

Ultimately, our work aims to move beyond raw performance scores toward a deeper understanding of generalization behavior in AI-text detection. While linguistic features alone do not capture the full complexity of generalization, they provide a valuable starting point for interpreting detector behavior in the wild.

\section{Related Work}

\paragraph{Datasets for AI Text Detection.}
Recent benchmarks have advanced AI-generated text detection by covering multiple languages \cite{wang-etal-2024-m4gt, macko_multisocial_2024}, domains \cite{li_mage_2024, verma_ghostbuster_2024, dugan-etal-2024-raid}, and generator models \cite{hu2023radar, abassy_llm-detectaive_2024, tao_cudrt_2024}. Several works also consider mixed-authorship settings \cite{yu2024cheatlargescaledatasetdetecting, 10.1007/978-3-031-57850-2_16}. However, few datasets jointly evaluate the impact of LLM type, domain, and prompt style in a systematic and controlled manner. Prompt engineering remains particularly underexplored, despite its known influence on generation behavior. To address these gaps, we introduce a new dataset that enables controlled experiments across 7 LLMs, 4 domains, and 6 prompting strategies.

\paragraph{AI Text Detection Models.}
Detection methods mainly fall into two broad categories: statistical detectors (e.g., GLTR \cite{gehrmann-etal-2019-gltr}, DetectGPT \cite{mitchell2023detectgptzeroshotmachinegeneratedtext}, Binoculars \cite{hans2024spotting}) and fine-tuned classifiers using pretrained LMs (e.g., RoBERTa \cite{liu2019robertarobustlyoptimizedbert}, DeBERTa \cite{he2021deberta}, XLM-R \cite{conneau-etal-2020-unsupervised}). While statistical detectors offer interpretability, classifier-based approaches consistently achieve stronger performance on benchmark datasets such as M4GT \cite{wang-etal-2024-m4gt}, MULTITuDE \cite{macko_multitude_2023}, and MultiSocial \cite{macko_multisocial_2024}. Our study builds on this foundation by fine-tuning two top-performing detectors, XLM-RoBERTa and DeBERTa-V3, across varied training settings to assess their generalization.

\paragraph{Generalization in Detection.}
Generalization is a core challenge in AI-text detection. Prior work has shown that detectors trained on one task or domain often fail when evaluated on others \cite{xu-etal-2024-generalization, li_mage_2024, bhattacharjee2024eagledomaingeneralizationframework}. However, most studies focus on reporting performance gaps without offering deeper explanations. Moreover, while some papers examine prompt-based variation, they typically limit prompting strategies or focus on handcrafted or adversarial prompts \cite{xu-etal-2024-generalization,10.1007/978-3-031-57850-2_16}. In contrast, our work evaluates generalization comprehensively across prompt styles, model families, and content domains, includes both naturalistic generation strategies (e.g., few-shot, CoT) and handcrafted prompts (1-shot CoT, self-refine).

\paragraph{Explaining Generalization Behavior.}
Several works have proposed high-level explanations for generalization variance. For example, \citet{xu-etal-2024-generalization} attribute success to \textit{prompt similarity} and \textit{human–LLM alignment}, while \citet{li_mage_2024} explore distributional differences using linguistic metrics like POS tags and named entity counts. However, these studies stop short of identifying which specific linguistic features correlate with generalization. Our work advances this line of inquiry through a detailed correlation analysis of 80 linguistic features, covering syntactic, stylistic, and discourse-level signals. We quantify feature shifts between training and testing data and link them to generalization performance, revealing interpretable signals, e.g., shifts in pronoun frequency or passive voice usage that influence detector robustness.

\section{Dataset Creation}

To provide a comprehensive study of generalization of AI-text detectors against prompt, LLM and dataset changes, we create our human-written and AI-generated text dataset by incorporating 6 prompting strategies, 7 LLMs with different parameter sizes from different model families, and 4 datasets from different domains. 

\subsection{Human-written text}
We first randomly sample the human-written text of different domains from  4 datasets: (1) Scientific paper \textbf{abstracts} from the arXiv dataset \citep{see-etal-2017-get}; (2) Product \textbf{reviews} from the AmazonReviews2023 dataset \citep{hou2024bridginglanguageitemsretrieval}; (3) \textbf{News} articles from the CNN/Daily Mail dataset \citep{clement2019usearxivdataset}; (4) Question and answers (\textbf{QA}) from the ASQA dataset \citep{stelmakh2023asqafactoidquestionsmeet}.

For abstracts and news articles, we use only texts that are at least 1,000 characters long. We set the minimum text length for reviews to 350 characters because the original text is short. For the QA dataset, we sample the longest texts considering the limited size of the data.
For each of the datasets, we sample 3,000 examples and split them into training, validation and testing set with a split ratio of 50:17:33.

\paragraph{Data cleaning for human-written text.} To remove obvious features for AI-text detectors, we use the following data cleaning steps for all the human-written texts: (1) Removing duplicates; (2) Normalizing punctuation; (3) Removing duplicated whitespace; (4) Removing URLs, e-mail addresses, and emojis; (5) Artifacts such as dates of article;  (6) Filtering non-English text; (7) Filter too short text, as text length can impact the difficulty of the task \cite{wang-etal-2024-m4}.




\subsection{AI Text Generated with LLMs}

The AI-text part of the dataset consists of texts generated by 7 LLMs using 6 diverse prompting strategies for each dataset. For each human-written text, we apply every LLM and prompting strategy to generate an AI‑text counterpart under the same topic. Consequently, for each source dataset, the final data include 3,000 human-written texts and, for every model–prompt combination, 3,000 corresponding AI-generated texts. In total, the dataset comprises 516,000 texts: 12,000 human-written and 504,000 AI-text. Each AI text is matched with its human-written text counterpart for performing a binary classification.

\paragraph{LLMs.} We employ LLMs with different parameter sizes and from 5 model families: (1) \textbf{Mistral 123B} \cite{mistral2024largeinstruct}: Mistral-Large-Instruct-2411; (2) \textbf{Deepseek 70B} \cite{deepseekai2025deepseekr1incentivizingreasoningcapability}: 
DeepSeek-R1-Distill-Llama-70B; (3) \textbf{Llama 70B} \cite{meta2024llama3}: Llama-3.3-70B-Instruct; (4) \textbf{Qwen 72B, Qwen 32B, Qwen 14B} \cite{qwen2024qwen2.5}: Qwen2.5-72B/-32B/-14B-Instruct; (5) \textbf{Solar 22B} \cite{upstage_solar_pro_preview_instruct_2024}: solar-pro-preview-instruct.

\paragraph{Prompts.} We use 6 different prompting strategies based on existing research on prompt engineering. The prompts include:
(1) \textbf{0-shot} prompts that only provide the metadata (e.g., title, text length) of each data sample; (2) \textbf{3-shot} prompts \cite{brown2020languagemodelsfewshotlearners} that contains 3 human-written texts from the same dataset for in-context learning; (3) \textbf{Style} prompts \cite{10.1007/978-3-031-57850-2_16} which require LLMs to write in a style like the given human-written text example; (4) \textbf{0-shot CoT} prompts \cite{kojima2023largelanguagemodelszeroshot} which consist of phrase ``let’s think step by
step.''; (5) \textbf{1-shot CoT} prompts \cite{wei2022chain} that contain manually written step-by-step instructions, the instruction is based on an example of a human-written text; and (6) \textbf{Self-refine} prompts \cite{madaan2023selfrefineiterativerefinementselffeedback} that use the LLM itself to critique and improve its own responses. Self-refine prompts are multi-stage prompts that comprise 4 stages: firstly, the LLM is prompted to generate the AI text, then it is requested to provide feedback on how to make the generated AI text more human-like. Later, the LLM needs to incorporate the feedback to improve the initially generated AI text. The improved text is final if the LLM judges it sounds more human-written than the human-written text counterpart; otherwise, it goes back to the feedback step for at most 3 iterations. For each prompting strategy, we modified the prompt template used for each dataset to suit the task of text generation. To match the text length of the AI-generated texts to their human-written counterparts, we include information about character count in the prompts.
A detailed discussion of the prompts can be found in the appendix \ref{prompt-appendix}, along with the prompt templates used to create our dataset (Table \ref{tab:promp-detail}).

\paragraph{Data Cleaning for AI-Generated Text.}  
To prevent detectors from exploiting superficial artifacts rather than genuine linguistic characteristics, we extensively clean the LLM-generated texts by removing elements that could trivially reveal their artificial origin. Specifically, we remove formulaic AI responses (e.g., \textit{``Certainly!''}, \textit{``Sure!''}), structural markers such as section titles, bullet points, and numbered lists, placeholder tokens in square brackets (e.g., \textit{[your name]}, \textit{[insert e-mail address]}), extraneous metadata including review ratings, character-count information, and sentences beginning with \textit{``Note:''} that describe the generation process, non-linguistic symbols such as asterisks (\texttt{*}), triple dashes (\texttt{---}), and hash symbols (\texttt{\#}), as well as model-specific tags such as \texttt{\textbackslash think} and any preceding text in Deepseek reasoning outputs. This cleaning step ensures that the evaluation focuses on the linguistic properties of the generated text rather than on easily detectable formatting artifacts.

\section{Generalization of AI-Text Detectors}

We introduce three evaluation settings to assess the generalization ability of AI-text detectors across three dimensions: prompts, LLMs, and datasets.

\subsection{Generalization Testing}

Assume an AI-text detector $M$ is fine-tuned on a training set $\mathcal{D}_{i,j,k}^{\text{train}}$, consisting of AI-generated texts produced with prompt $p_i$ by a generative model $g_j$ for dataset $d_k$, along with human-written texts. We evaluate the cross-prompt, cross-model, and cross-dataset generalization of the AI-text detector under the following settings.

\paragraph{Cross-prompt (C-P) testing.}  
This setting evaluates how well detectors generalize to texts generated with prompting strategies unseen during training. Each detector is evaluated on test sets produced by the same LLM and drawn from the same dataset, but generated with different prompts.  This controlled setting ensures that the only changing factor is prompt strategies during evaluation.
For example, a detector trained on the training split of QA texts generated by Llama 70B with 0-shot prompts is evaluated on the test split generated by Llama 70B with all 6 prompt types. Formally, the accuracy of the detector on the test data $D^{test}$ when trained on $D^{train}$ is denoted as $\text{Acc}(M(D^{test}|D^{train}))$. Thus, the generalization accuracies from prompt $p_i$ to all prompts are formalized as a list:
\begin{equation}
   \Delta_{\text{gen}}^{(p_i)} = \left\{\text{Acc}\!\left(M\!\left(\mathcal{D}_{c,j,k}^{\text{test}} \mid \mathcal{D}_{i,j,k}^{\text{train}}\right)\right)\right\}_{c=1}^6.
\end{equation}

Where $\Delta_{\text{gen}}^{(p_i)}$ is a list of 6 accuracy values, with each value representing the generalization accuracy of a prompt. We carried out the test for each prompt and resulted in a 2-dimensional 6x6 vector (plot like left heatmaps in Figure \ref{fig:cross-prompt-spec}),  which presents the cross-prompt result of all prompts in one of the conditions, denoted as $\Delta_{\text{gen}}^{(p)}$.

\paragraph{Cross-model (C-M) testing.}  
This setting evaluates generalization to texts generated by LLMs not seen during training. A detector fine-tuned on the training split from one LLM is evaluated on the test splits produced by other LLMs.  
For example, a detector trained on abstracts generated by Llama 70B is tested on abstracts generated by all 7 LLMs. We use 0-shot prompts ($p_1$) in this setting. Formally:
\begin{equation}
   \Delta_{\text{gen}}^{(g_j)} = \left\{\text{Acc}\!\left(M\!\left(\mathcal{D}_{1,c,k}^{\text{test}} \mid \mathcal{D}_{1,j,k}^{\text{train}}\right)\right)\right\}_{c=1}^7.
\end{equation}

Similar to cross-prompt testing, we apply the formula to all LLMs, obtaining in a 7x7 vector as cross-model results $\Delta_{\text{gen}}^{(g)}$.

\paragraph{Cross-dataset (C-D) testing.}  
This setting evaluates generalization across different dataset domains. A detector fine-tuned on the training split from one dataset is evaluated on test splits from other datasets generated by the same LLM.  
For example, a detector trained on abstracts generated by Llama 70B is tested on news, reviews, and QA data generated by the same LLM. We use 0-shot prompts ($p_1$) in this setting. Formally:
\begin{equation}
   \Delta_{\text{gen}}^{(d_k)} = \left\{\text{Acc}\!\left(M\!\left(\mathcal{D}_{1,j,c}^{\text{test}} \mid \mathcal{D}_{1,j,k}^{\text{train}}\right)\right)\right\}_{c=1}^4.
\end{equation}

The corresponding cross-dataset result $\Delta_{\text{gen}}^{(d)}$ is a 4x4 vector.

\subsection{Training setup}

 We fine-tune XLM-RoBERTa-base \citep{conneau-etal-2020-unsupervised} (referred to as RoBERTa) and DeBERTa-V3-small \citep{he2021debertav3} (referred to as DeBERTa) for binary classification to distinguish between human-written and AI-generated text. These architectures have achieved state-of-the-art performance in prior work \citep{wang-etal-2024-m4gt}.
 We train a separate detector for each combination of prompt type, AI-text generation model, and dataset type, resulting in 168 (i.e., 7x4x6) in-domain detectors when using, for example, RoBERTa for fine-tuning. 

 Model fine-tuning is performed using the following hyperparameters: \texttt{learning rate} = 2e-5, \texttt{num train epochs} = 3, \texttt{weight decay} = 0.01, \texttt{train batch size} = 16.
The maximum sequence length is set to 512, corresponding to the maximum input length supported by XLM-RoBERTa. All our experiments are conducted on
NVIDIA HGX H100, and approximately 400 GPU hours to replicate.

\section{Explaining Generalization with Linguistic Analysis}

To better understand what causes the variance of generalization performance, we perform a comprehensive linguistic feature analysis by measuring the correlation of 80 different feature metrics with the generation results. 
We first introduce the definition and metric of each linguistic feature and present the correlation evaluation method.

\subsection{Linguistic Feature Definitions and Metrics}
Our studied features can be categorized into Lexical diversity, Lexical density, Sentiment, Readability, Part-of-Speech (POS), and Grammatical and Lexical analysis. We introduce the most correlated features and metrics in the main paper and the rest in Appendix \ref{app:linguistic}.

\paragraph{Readability.} Readability refers to the ease of understanding a text. AI-generated text tends to be less readable than human-written text \cite{doi:10.1177/0261927X231200201,10545131}. We use the \textbf{Gunning fog index} \cite{yadagiri-etal-2024-detecting} as one of the measures of readability, which is an estimated number of years needed to understand a given passage.

\paragraph{Part-of-Speech (POS).} We use a selection of metrics from StyloMetrix \cite{okulska2023stylometrixopensourcemultilingualtool} to compare the frequency of \textbf{verbs, nouns, adjectives, numerals, etc}. The frequency of parts of speech is measured as the fraction of text covered by tokens representing a given part of speech. 
Previous research has discovered differences between human-written and AI-generated text in terms of frequency of certain POS \cite{georgiou2024differentiatinghumanwrittenaigeneratedtexts}. Therefore, POS analysis is relevant for gaining insights into the literary style of texts in our dataset.
\begin{figure*}[h!]
    \centering
    \begin{subfigure}[b]{1\textwidth}
    \includegraphics[width=1\linewidth]{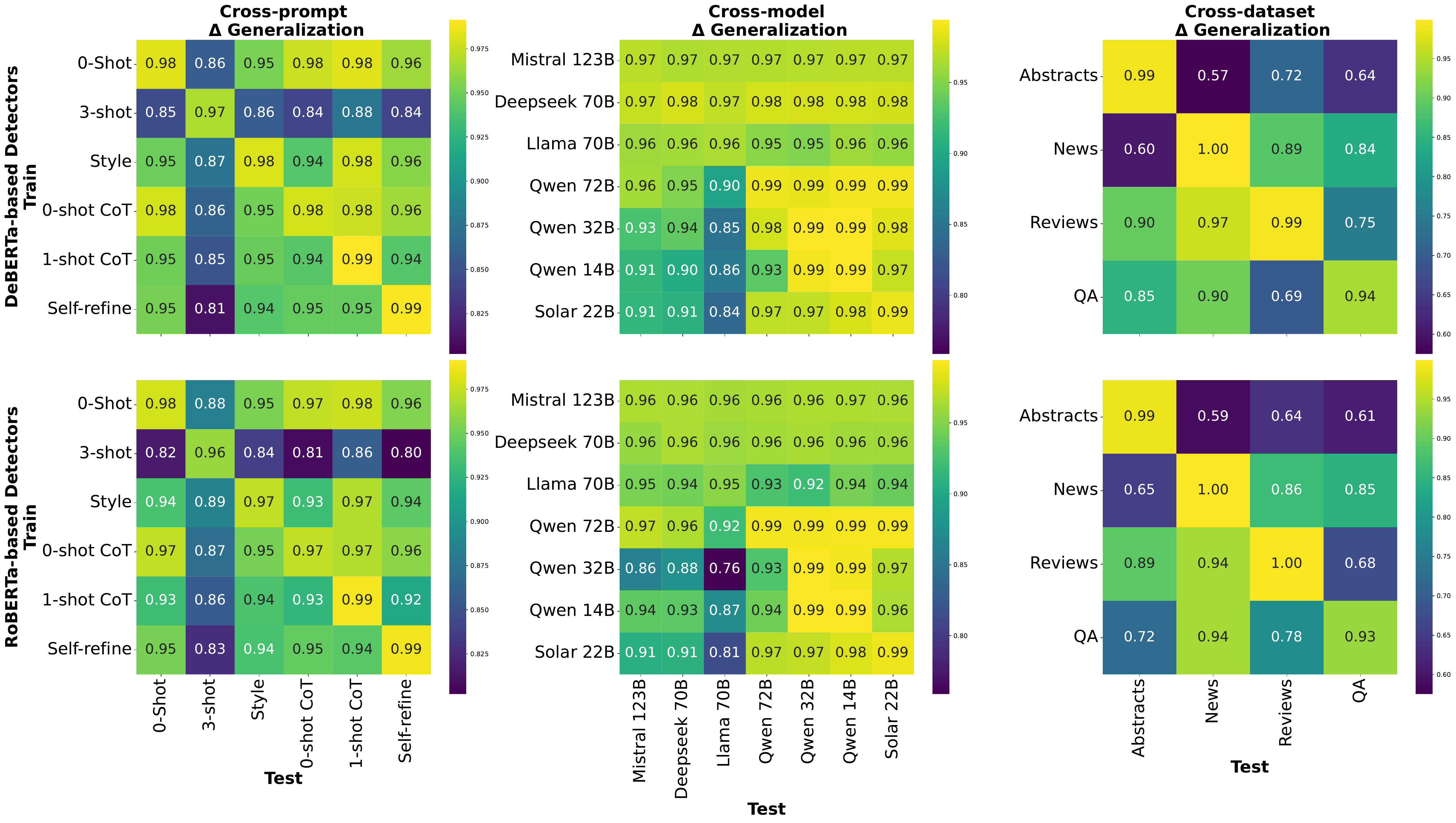}
     \caption{The aggregated generalization performance.}\label{fig:avg_performance_heatmap}
    \end{subfigure}
    \begin{subfigure}[b]{1\textwidth}
    \includegraphics[width=1\linewidth]{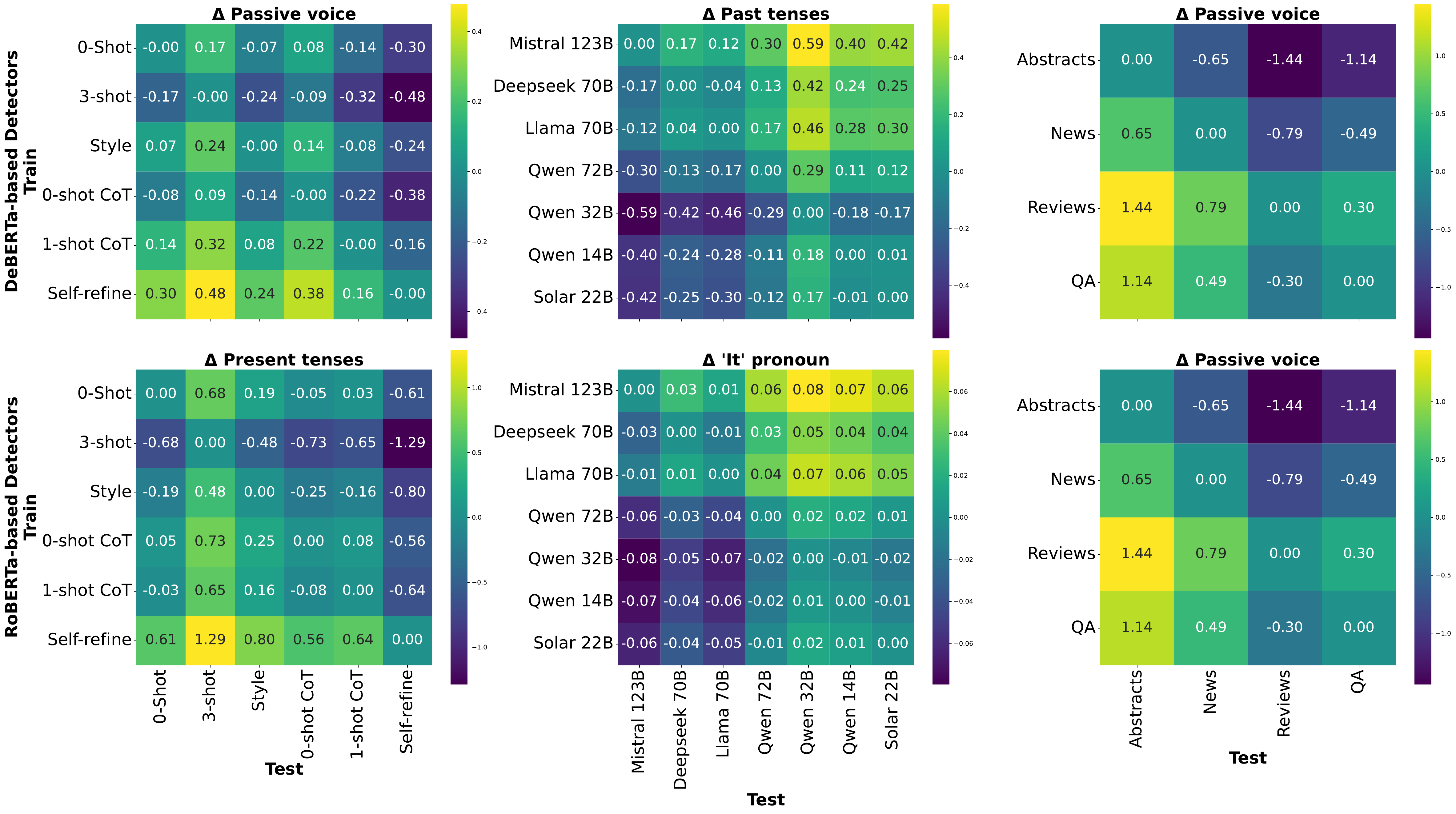}
    \caption{The aggregated feature shifts between training and test configurations.}\label{fig:avg_feature_heatmap}
    \end{subfigure}
    \caption{Comparison of aggregated generalization performance and aggregated feature shifts across all evaluation settings. Similar patterns in the two heatmaps indicate that certain feature shifts are correlated with reduced generalization accuracy.}
    \label{fig:main-corr-heatmap}
\end{figure*}

\paragraph{Grammatical Analysis.} We use StyloMetrix \cite{okulska2023stylometrixopensourcemultilingualtool} metrics to compare texts in terms of grammatical categories related to verbs. Human-written texts have been shown to contain more passive voice than AI-generated texts \cite{georgiou2024differentiatinghumanwrittenaigeneratedtexts}. We measure the \textbf{incidence of passive and active voice} as the frequency of verbs in passive or active voice. We also compare the differences between the choice of tenses. The \textbf{frequency of past, present and future tenses} is measured as the fraction of the text covered by verbs in past, present and future tenses. For example, the incidence of past tenses is the number of verbs in past simple, past continuous, past perfect or past perfect continuous divided by the total number of words in the text.

\paragraph{Lexical Analysis.} As part of lexical analysis, we measure the \textbf{frequency of personal names}, and  \textbf{adjectives in comparative and superlative degrees}. The pronoun-related metrics \cite{okulska2023stylometrixopensourcemultilingualtool} analyze the differences in the usage of pronouns in human-written and AI-generated text. We calculate the frequency of specific personal or reflexive pronouns and certain types of pronouns (e.g., ``We'', ``It'', ``Our'', ``Yourself''). 

\subsection{Correlation Between Generalization and Linguistic Features}

To further investigate factors that influence generalization, we examine how changes in linguistic features correlate with generalization performance. For each linguistic feature $f$, we compute its shift between a given training configuration and the corresponding test configuration:
\begin{equation}
\Delta_f = f(\mathcal{D}^{\text{train}}) - f(\mathcal{D}^{\text{test}})
\end{equation}
where $f(\mathcal{D}^{\text{train}})$ denotes the feature difference between AI-generated and human-written texts in the training configuration ($f(\mathcal{D}^{\text{train}}) = f(\mathcal{D}^{\text{train}}_{human}) - f(\mathcal{D}^{\text{train}}_{AI})$), and $f(\mathcal{D}^{\text{test}})$ denotes the feature difference in the corresponding test configuration. Corresponding to the generalization testing, we denote the cross-prompt, cross-model, and cross-dataset feature shift as $\Delta^{(p)}_f$, $\Delta^{(g)}_f$, and $\Delta^{(d)}_f$, respectively, which all have the same size. 

We then compute the Pearson correlation between \textbf{flattened} generalization accuracy and these feature shifts under the same conditions:
\begin{equation}
Corr(f) = |\text{Pearson}\big(\Delta_{\text{gen}}^{(n)}, \Delta_f^{(n)}\big)|
\end{equation}

where $n$ indexes each cross-prompt, cross-model, or cross-dataset comparison. The resulting value $Corr(f)$ lies in $[0,1]$, with $0.1 \leq Corr(f)<0.3$ indicating a low correlation, $0.3 \leq Corr(f) < 0.5$,  $0.5 \leq Corr(f) < 0.7$ and $Corr(f) \geq 0.7$ as  moderate,  high  and  strong correlations \cite{datatab2025pearson}. This analysis allows us to identify which linguistic features are most strongly associated with robust generalization across different test settings.

\textbf{Setting-specific correlation} (Table \ref{tab:cross-prompt-sep-corr}) is measured under a specific testing combination. For example, the setting-specific correlation of cross-prompt generalization and linguistic feature shifts is only measured on texts from Llama 70B and the Abstract dataset.

\textbf{Overall correlation} (Table \ref{tab:main-corr}) is measured under all combinations. For example, the overall correlation of cross-prompt generalization and linguistic feature shifts is measured on the texts of every combination of LLMs and datasets.

\input{tables/heatmap_corr_table}
\input{tables/prompt_sep_corr_table}

\textbf{Aggregation of Results.}  
To provide a high-level summary of cross-prompt, cross-model, and cross-dataset generalization and feature shifts (as Figure \ref{fig:avg_performance_heatmap} and \ref{fig:avg_feature_heatmap}), we report accuracy and shift values averaged over the dimensions that are not the focus of the evaluation. For instance, when presenting overall cross-prompt results, we average accuracy scores across all 7 LLMs and 4 datasets.

\section{Results and Analysis}

We analyze the results in Table~\ref{tab:main-corr}, Table~\ref{tab:cross-prompt-sep-corr}, and Figure~\ref{fig:main-corr-heatmap} to understand how generalization performance is shaped by shifts in linguistic features.

\subsection{General Findings}
\textbf{Finding 1: Linguistic features have significant correlations with generalization results.}  
Table~\ref{tab:main-corr} shows that several features exhibit significant correlations (bold values) with detector generalization. For example, overall \textit{cross‑model generalization} that averaged across datasets is moderately correlated (0.416) with the proportion of past‑tense verbs, indicating that stylistic verb usage in training data influences transfer. In contrast, cross‑prompt generalization averaged across datasets and LLMs shows only weak correlation with passive voice. These results highlight that some features play a more critical role than others in determining generalization success.

\textbf{Finding 2: Certain dataset–model combinations reveal {very strong} feature dependencies.}  
Although overall cross‑prompt correlations appear weak in Table~\ref{tab:main-corr}, Table~\ref{tab:cross-prompt-sep-corr} reveals that in specific configurations the effect is dramatic. For instance, on Llama‑70B outputs for the Reviews dataset, cross‑prompt generalization is strongly correlated (>$0.7$) with the number of short sentences. Fig~\ref{fig:llama-reviews} shows that changes in this feature align directly with sharp drops in performance when generalizing from 1‑shot CoT to other prompting strategies. This demonstrates that some detectors are highly sensitive to prompt‑induced shifts in linguistic structure.

\textbf{Finding 3: Different detectors rely on different linguistic features.}  
RoBERTa‑based detectors and DeBERTa‑based detectors do not exploit the same linguistic signals. For example, the cross‑model generalization of RoBERTa models is moderately correlated (0.385) with the frequency of ``It'' pronouns, whereas for DeBERTa the correlation is only 0.236. This suggests that detectors may learn fundamentally different features for distinguishing human and AI text even when trained on the same data.

\subsection{Linguistic Analysis of Generalization Results}

Figure~\ref{fig:main-corr} summarizes average performance across the three generalization settings. As expected, \textbf{in‑domain testing achieves near‑perfect accuracy}, as shown in the diagonal cells of Figure~\ref{fig:avg_performance_heatmap}, confirming the strong baseline capabilities of our detectors. Detailed in‑domain results are reported in Table~\ref{tab:results} in the Appendix.

\subsubsection{Cross‑prompt Generalization}
The most striking pattern is that the \textbf{3‑shot prompt is consistently the hardest to generalize to and from}, with accuracy dropping to 80–89\%. Other prompting strategies show relatively minor effects.

\begin{figure}[h]
    \centering
    \begin{subfigure}[b]{0.48\textwidth}
    \includegraphics[width=1\linewidth]{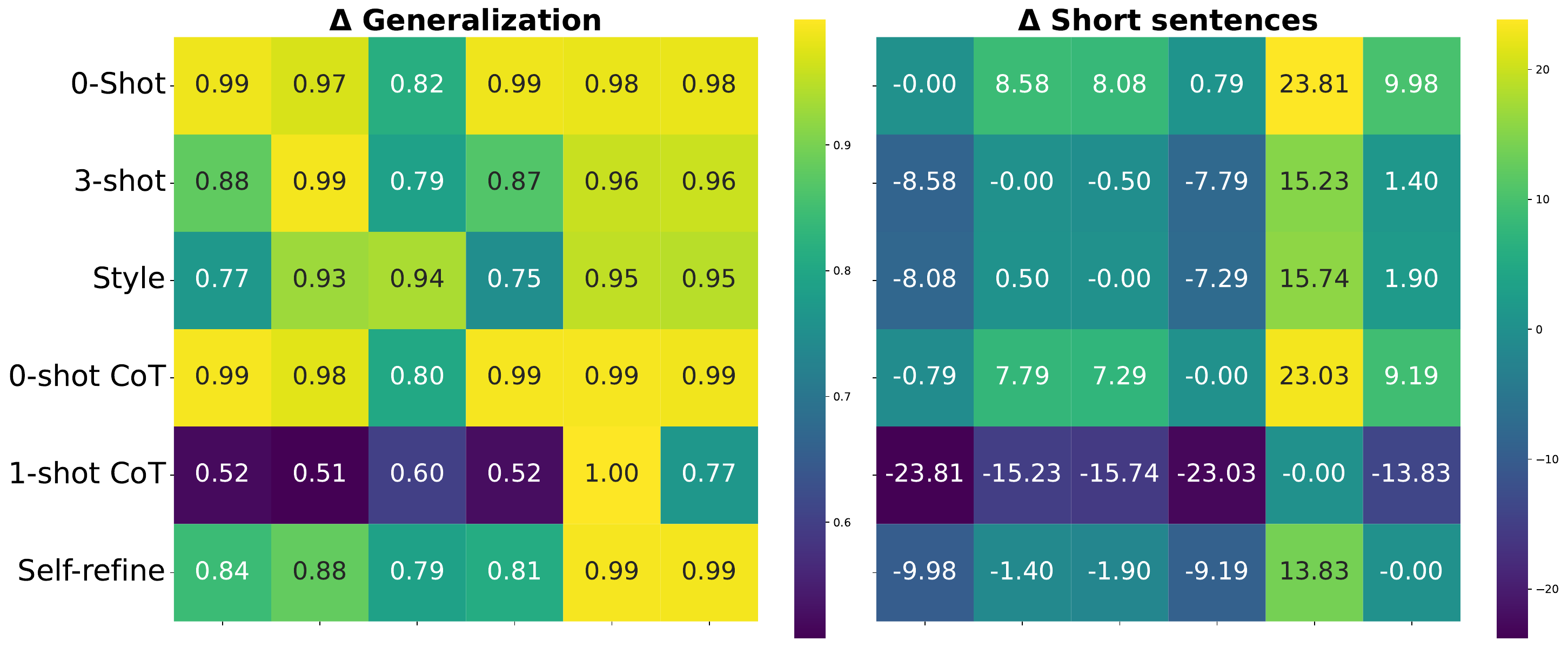}
    \caption{Llama 70B, Reviews}\label{fig:llama-reviews}
    \end{subfigure}
    \begin{subfigure}[b]{0.48\textwidth}
    \includegraphics[width=1\linewidth]{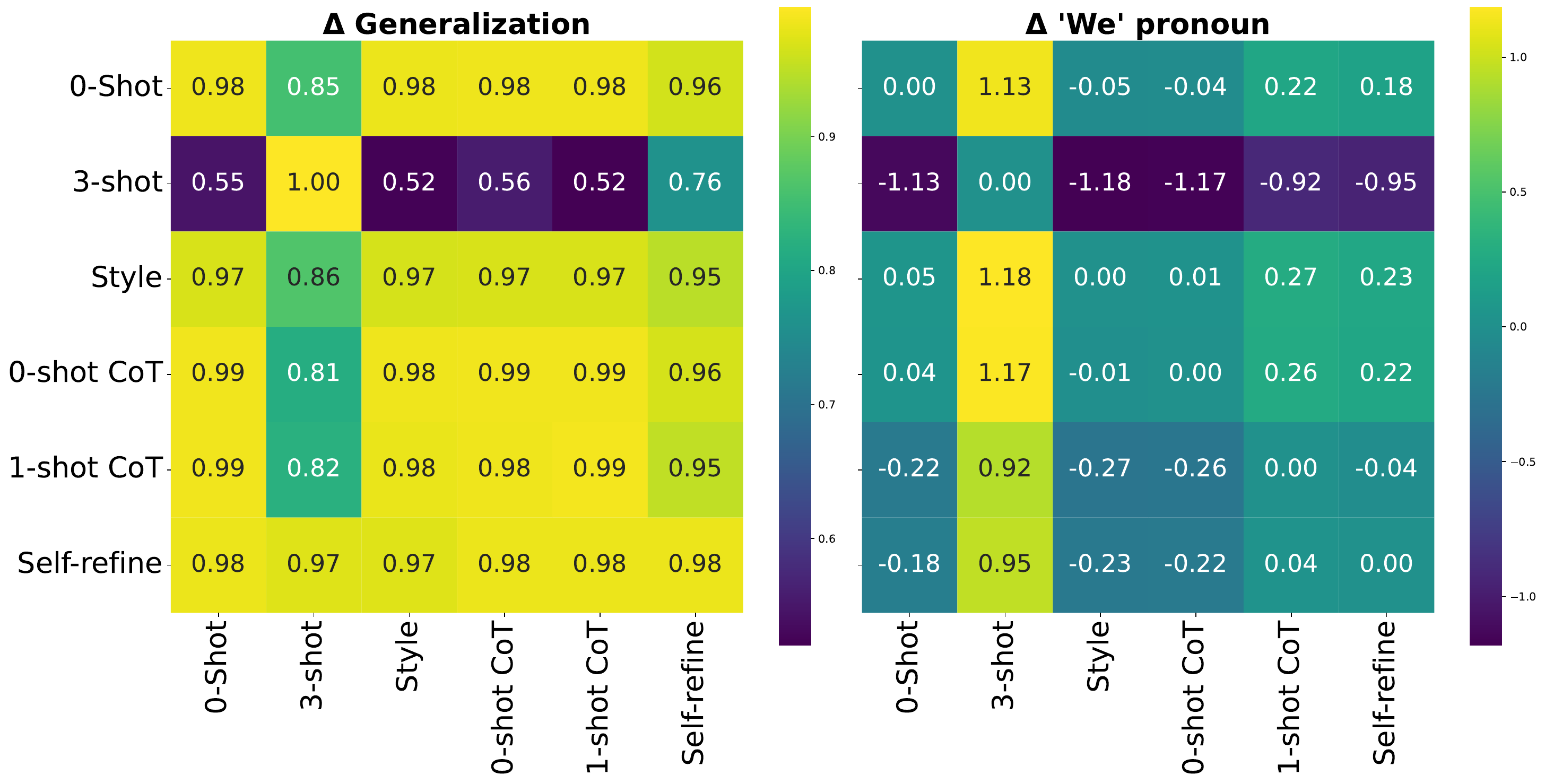}
    \caption{Deepseek 70B, Abstracts}\label{fig:deepseek-abstract}
    \end{subfigure}
    \caption{Cross-prompt generalization and feature shifts when evaluating on a specific model and dataset. A  clearer and stronger correlation is observed than the overall cross-prompt correlation in Figure \ref{fig:main-corr-heatmap}. We also show the specific case study of the generated texts for the above two settings in Table \ref{tab:case_study_a} and \ref{tab:case_study_b} in the Appendix.}
    \label{fig:cross-prompt-spec}
\end{figure}

\textbf{Explanation.}  
Figure~\ref{fig:avg_feature_heatmap} shows averaged feature shifts, but strong effects can be masked by aggregation. For example, Figure~\ref{fig:deepseek-abstract} highlights a clear pattern: AI texts that use the ``We'' pronoun in similar contexts are more difficult to generalize to for the detectors. This confirms Findings~2 and~3: when we zoom in on specific dataset–model pairs, clear linguistic drivers of generalization emerge.

\subsubsection{Cross‑model Generalization}
A major finding is that \textbf{detectors trained on Qwen or Solar outputs perform poorly on Llama‑generated text}, whereas generalization across other LLMs is more stable.

\textbf{Explanation.}  
Figures~\ref{fig:avg_performance_heatmap} and~\ref{fig:avg_feature_heatmap} reveal that cross‑model generalization is {moderately influenced by shifts in past‑tense usage and ``It'' pronoun frequency}. Qwen and Solar outputs share similar linguistic profiles, which diverge from Llama’s, explaining this degradation.

\subsubsection{Cross‑dataset Generalization}
The most pronounced performance gap appears here: \textbf{detectors trained on abstracts generalize poorly to other datasets, achieving as low as 57\% accuracy when tested on news articles}. Conversely, detectors trained on reviews or QA data transfer more successfully to news.

\textbf{Explanation.}  
Across all detectors, cross‑dataset generalization shows {moderate correlation with passive voice usage}. Abstracts exhibit a higher rate of passive constructions, which likely makes them a poor source domain for training detectors that must generalize broadly.

\begin{table*}[t]
\centering
\footnotesize
\begin{tabular}{lccccccc}
\toprule
& & \multicolumn{3}{c}{\textbf{DeBERTa-based }} & \multicolumn{3}{c}{\textbf{RoBERTa-based}} \\
\cmidrule(lr){3-5}\cmidrule(lr){6-8}
 && C-P & C-M & C-D & C-P & C-M & C-D \\
\midrule

\multirow{5}{*}{\textbf{Pearson}} & Bonferroni  & 2 & 15 & 0 & 3 & 28 & 0 \\
&FDR         & 2 & 37 & 0 & 6 & 58 & 0 \\
\cmidrule{2-8}
&\multirow{3}{*}{Top features} & Numerals &Past tense &\multirow{3}{*}{--} &``Our'' pronoun &``It'' pronoun& \multirow{3}{*}{--} \\
& &  Passive voice &Personal names & &Present tense &Active voice & \\
& &  &Numerals & &``We'' pronoun &``Yourself'' pronoun & \\
\midrule
\multirow{5}{*}{\textbf{Spearman}} & Bonferroni & 3 & 16 & 0 & 2 & 25 & 0 \\
& FDR         & 4 & 44 & 0 & 10 & 51 & 0 \\
\cmidrule{2-8}
&\multirow{3}{*}{Top features} &``She'' pronoun &MATTR &\multirow{3}{*}{--} &``She'' pronoun &Active voice&\multirow{3}{*}{--}  \\
& & ``He'' pronoun &FLESCH & &``Her'' pronoun &``It'' pronoun & \\
& &short sentences & Gunning Fog & &Present tense &Function words & \\
\bottomrule
\end{tabular}
\caption{Robustness study of applying multiple-hypothesis correction to Pearson (linear) and Spearman (non-linear) correlation. The numerical values represent the number of features that remain significant after the multiple-hypothesis correction.}\label{robust_stuudy}
\end{table*}

\subsection{Robustness Study}
 We conduct the robustness study using multiple-hypothesis corrections (Bonferroni \cite{goeman2014multiple} and Benjamini–Hochberg FDR \cite{bogdan2008comparison}), and further including Spearman (non-linear) correlations \cite{ali2022spearman}. The results are shown in Table \ref{robust_stuudy}, we find that:

(1) Key linguistic correlates (pronoun usage, verb tense, active/passive voice) remain robust for cross-prompt and cross-model generalization.

(2) A substantial number of features (>=15) remain significant in cross-model settings across various settings.

(3) Non-linear effects emerge under Spearman correlations, strengthening our interpretation of detector behavior.

(4) For cross-dataset generalization, no individual features remain significant under strict multiple-hypothesis correction, suggesting that performance degradation is likely driven by broader distributional shifts rather than a single dominant linguistic cue.

Importantly, our main conclusions remain valid after these robustness checks. 

\subsection{Discussion}

While our analysis highlights the role of linguistic features in explaining the generalization behavior of AI-text detectors, we acknowledge that these features represent only one facet of a more complex landscape. Generalization performance is likely influenced by a broader set of factors, including semantic coherence, discourse structure, and detector-specific inductive biases. Our findings should therefore be interpreted as offering a linguistic perspective rather than a comprehensive account of generalization. Nonetheless, by systematically correlating linguistic feature shifts with detection performance, our study contributes valuable insights into how stylistic and grammatical signals may impact detector robustness across prompts, models, and domains.

\section{Conclusion}

This work presents an interpretable investigation into the generalization behavior of AI-text detectors. While prior studies primarily report detection performance, we go further by examining why generalization succeeds or fails through a linguistic lens. Across a large-scale benchmark incorporating diverse prompts, LLMs, and domains, we show that state-of-the-art detectors, despite near-perfect in-domain accuracy, often struggle in cross-prompt, cross-model, and cross-dataset scenarios. To explain these generalization behaviors, we quantify shifts in 80 linguistic features between training and testing distributions and uncover statistically significant correlations between feature shifts and generalization performance.

Our analysis reveals that features such as pronoun usage, verb tense, and passive voice are predictive of generalization gaps, but their influence varies across detectors and settings. This suggests that detectors latch onto different linguistic signals depending on their training context, impacting their robustness in different testing scenarios.

These findings underscore two key points: (1) evaluation must extend beyond in-domain testing to realistically assess detector reliability, and (2) linguistic analysis provides a principled and interpretable path toward diagnosing and improving generalization.

\section*{Limitations}
While our work offers new insights into the generalization behavior of AI-text detectors through linguistic analysis, several limitations remain.

First, our study focuses on English-language text and detectors trained on English corpora. Although our methodology can be extended to multilingual settings, the linguistic features and generalization patterns may differ significantly across languages due to variations in grammar and stylistic conventions.

Second, we rely on surface-level linguistic features (e.g., POS tags, sentence length, voice, pronouns) that can be extracted using standard NLP tools. While these features provide interpretable signals, they may not capture deeper semantic or discourse-level properties that also influence detector decisions.

Third, the detectors we evaluate are based on fine-tuned encoder-only transformer models. Other architectures, such as generative or retrieval-augmented models, may exhibit different generalization behaviors and rely on alternative linguistic features.

Fourth, our correlation-based analysis reveals associations but does not establish causal relationships between feature shifts and performance drops. Further research using controlled interventions or counterfactual examples would be needed to verify causality.

Lastly, our dataset covers a wide but still limited set of domains, models, and prompting strategies. As the landscape of LLMs and prompting methods continues to evolve, future work should assess whether our findings hold for more recent or unseen generation techniques.

Despite these limitations, our study provides a strong foundation for more principled and interpretable evaluations of generalization in AI-text detection.

\section*{Acknowledgments} 
This research has been funded by the Vienna Science and Technology Fund (WWTF)[10.47379/VRG19008] ``Knowledge infused Deep Learning for Natural Language Processing''.
\bibliography{custom}
\appendix
\section{Appendix}
\label{sec:appendix}

\input{prompt_detail}

\subsection{Prompt templates}
We demonstrate detailed prompts used in the paper in Table \ref{tab:promp-detail}.

\subsection{Detailed classification results}

 \normalsize The results of the detailed in-domain accuracy are shown in Table \ref{tab:results}.


\subsection{Linguistic analysis}\label{app:linguistic}
\label{sec:linguistic_analysis}
\input{linguistic_analysis_appendix}

\input{tables/prompt_templates}

\input{tables/main_results_table}
\input{tables/corr_apen_table}

\begin{figure*}[htbp]
  \centering
  \begin{subfigure}[b]{1\textwidth}
    \centering
    \includegraphics[width=1\textwidth]{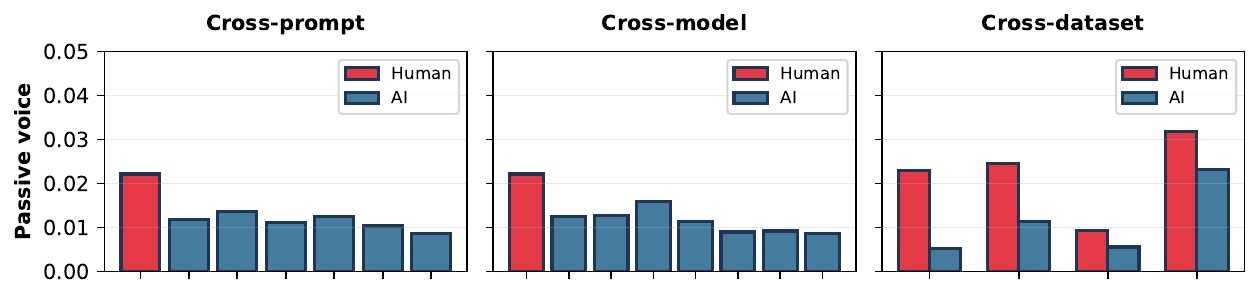}
    \caption{}
    \label{fig:main-corr-a}
  \end{subfigure}
  \begin{subfigure}[b]{1\textwidth}
    \centering
    \includegraphics[width=\textwidth]{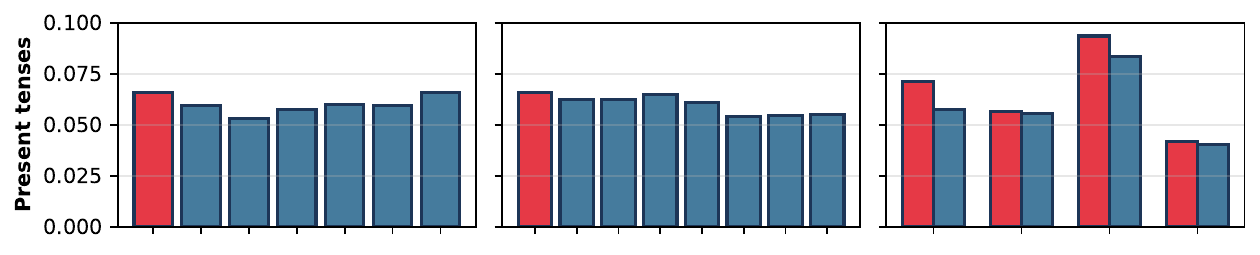}
    \caption{}
    \label{fig:main-corr-b}
  \end{subfigure}
    \begin{subfigure}[b]{1\textwidth}
    \centering
    \includegraphics[width=\textwidth]{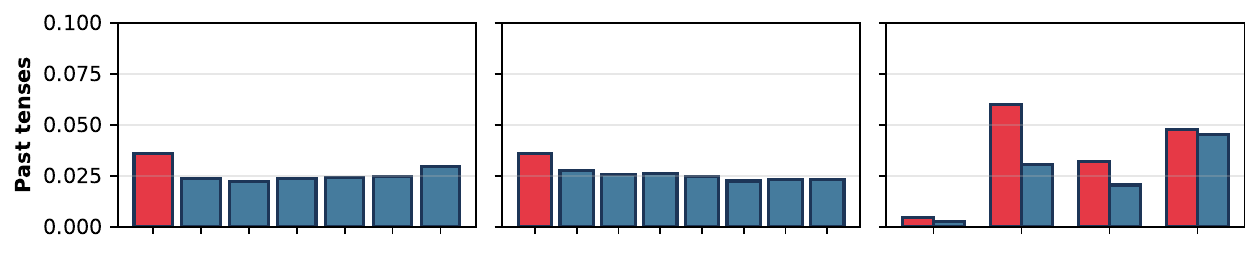}
    \caption{}
    \label{fig:main-corr-c}
  \end{subfigure}
    \begin{subfigure}[b]{1\textwidth}
    \centering
    \includegraphics[width=\textwidth]{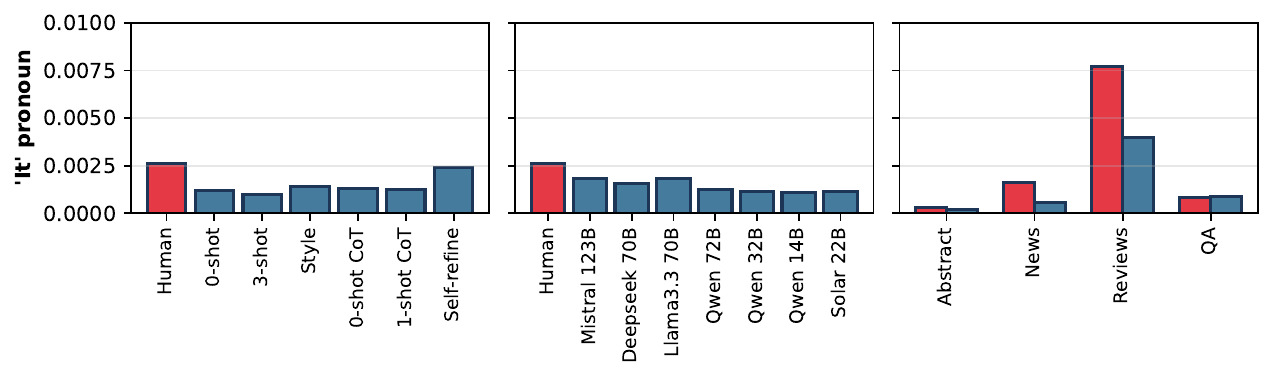}
    \caption{}
    \label{fig:main-corr-d}
  \end{subfigure}
  \caption{The more detailed comparison of different linguistic features across different configurations as well as the differences between human and AI text. We present the features that have the strongest correlations for different dimensions (underlined features in Table \ref{fig:main-corr} in the main paper).}
  \label{fig:main-corr}
\end{figure*}

\begin{table*}

    \centering
    \scriptsize
    \resizebox{\textwidth}{!}{\begin{tabular}{p{15.35cm}}
    \toprule
        \textbf{Dataset: Reviews}\\
     \textbf{Item name}: Homasy Amplified HDTV Antenna Indoor with Amplifier - 50 Mile Range, Built in Amplifier Signal Booster \\
    \textbf{Title}: Works Well\\
    \textbf{Rating}: 5\\ 
    \textbf{Review} length in characters: 859 \\
    \midrule
    
    \textbf{Human, Short sentences: 5 }\\
 \midrule
    We got rid of satellite TV a couple of years ago. We have two TVs in the house, but only one HD over the air tuner. We're looking to get another tuner, so with a second tuner, we need a second antenna. We tested this antenna with our existing tuner. It requires a USB port to be plugged in. We didn't attach it to our wall since we plan to use it in a different room and this was just a test. My husband held it up and was able to get a clear picture with it. Our living room has a two story ceiling. Our other antenna is up pretty high on the wall. The higher it was, the better the reception. The antenna is comparable to our other one in quality and performance. Now that we have another good antenna, we will go forward with purchasing a new tuner. I received this for free in exchange for my honest \& unbiased review. My thoughts \& opinions are my own.
\\
    \midrule
     \textbf{Llama70B, 0-shot, Short sentences: 2}\\\midrule
I'm impressed with this antenna! It was easy to set up and I get great reception on all my local channels. The amplifier really makes a difference, picking up stations that my old antenna couldn't. I live about 30 miles from the broadcast towers and it works flawlessly. Highly recommend!\\
 \midrule
 \textbf{Llama70B, 3-shot,  Short sentences: 1}\\\midrule
     I was a bit skeptical about buying an indoor antenna, but I'm glad I took the chance. The Homasy Amplified HDTV Antenna has exceeded my expectations. Setup was easy, just plug it in and scan for channels. I was able to receive over 30 channels, including major networks like ABC, CBS, NBC, FOX, and PBS, as well as some local stations. The picture quality is excellent, with no pixelation or freezing. The amplifier seems to make a big difference, as I was able to pick up channels that my old non-amplified antenna couldn't. The range is also impressive, I'm about 40 miles from the broadcast towers and I get a strong signal. The antenna is compact and lightweight, making it easy to place anywhere in the room. Overall, I'm very happy with this purchase and would recommend it to anyone looking for a reliable and affordable indoor HDTV antenna. Pros: Easy setup Excellent picture quality Good range (up to 50 miles) Compact and lightweight design Affordable price Cons: None that I can think of! Note: As with any antenna, the number of channels you'll receive will depend on your location and the strength of the signals in your area. But overall, I'm very satisfied with this product and would definitely recommend it.\\
     \midrule
     \textbf{Llama70B, Style, Short sentences: 2}\\\midrule
I bought the "\'Homasy Amplified HDTV Antenna Indoor with Amplifier"\' for my 75 year old dad because he likes to watch his local news and sports channels without relying on cable. This antenna makes it so he can. The thing I don\'t like? That you have to deal with all the extra cables the amplifier and signal booster are great, but they add clutter. So it would be nice to be able to just have a more streamlined setup. The extra cables just get tangled up behind the TV.\\
     \midrule
 \textbf{Llama70B, 0-shot CoT, Short sentences: 2}\\\midrule
Works Well I'm impressed with this antenna, it works well and picks up all local channels clearly. The amplifier helps boost weak signals, giving me more channel options. Easy to set up, compact design, and great picture quality. I can receive channels within a 50-mile range without any issues. Highly recommended!\\
\midrule
\textbf{Llama70B, 1-shot CoT, Short sentences: 0}\\\midrule
"Excellent Reception at a Great Price" I was a bit skeptical about purchasing this antenna due to mixed reviews, but it has exceeded my expectations in every way; it\'s a fantastic value for the money. The pros definitely outweigh the cons, as it picks up a ton of channels with crystal clear reception, and the amplifier makes a huge difference, allowing me to receive stations from over 40 miles away with perfect clarity. Yes, you may need to adjust the antenna\'s position for optimal signal strength, but that\'s a small trade-off for the quality and number of channels you get, and the fact that it\'s so easy to set up and use is just icing on the cake.
\\
\midrule 
     \textbf{Llama70B, self-refine, Short sentences: 0} \\\midrule
     I was a bit skeptical about whether this antenna would be able to pull in clear signals from 30 miles away, but I've been blown away by its performance. Setting it up was a breeze, and the built-in amplifier has made a huge difference in terms of picture quality I can now get all my local channels in crystal-clear HD, which is a big upgrade from the fuzzy reception I was getting before. What really impresses me, though, is that it's picked up a few channels I didn't even know existed, including some great local stations that I'd never been able to tune into before. Overall, I'm thoroughly impressed with this antenna and would definitely recommend it to anyone looking to boost their TV reception!\\
    \bottomrule
    \end{tabular}}
    \caption{Case study of human and AI texts for the cross-prompt generalization setting shown in Figure \ref{fig:cross-prompt-spec} (a).\\}\label{tab:case_study_a}
\end{table*}

\begin{table*}
    \centering
    \scriptsize
    \resizebox{\textwidth}{!}{\begin{tabular}{p{15.35cm}}
    \toprule
        \textbf{Dataset: Abstracts}\\
   
    \textbf{Title}: Satellites and large doping- and temperature-dependence of electronic
properties in hole-doped BaFe2As2\\
    \textbf{Category}: Strongly Correlated Electrons\\ 
    \textbf{Review length in characters}: 1037 \\
    \midrule
    
    \textbf{Human, ``We'' pronoun: 0.013698}\\
 \midrule
    Over the last years, superconductivity has been discovered in several families of iron-based compounds. Despite intense research, even basic electronic properties of these materials, such as Fermi surfaces, effective electron masses, or orbital characters are still subject to debate. Here, we address an issue that has not been considered before, namely the consequences of dynamical screening of the Coulomb interactions among Fe-d electrons. We demonstrate its importance not only for correlation satellites seen in photoemission spectroscopy, but also for the low-energy electronic structure. From our analysis of the normal phase of BaFe2As2 emerges the picture of a strongly correlated compound with strongly doping- and temperature-dependent properties. In the hole overdoped regime, an incoherent metal is found, while Fermi-liquid behavior is recovered in the undoped compound. At optimal doping, the self-energy exhibits an unusual square-root energy dependence which leads to strong band renormalizations near the Fermi level.
\\
    \midrule
     \textbf{Deepseek, 0-shot, ``We'' pronoun: 0.011628}\\\midrule
This study investigates the electronic properties of hole-doped BaFe₂As₂ using angle-resolved photoemission spectroscopy (ARPES). We observe prominent satellite bands whose intensity varies significantly with doping and temperature. These satellites reveal intricate many-body interactions, suggesting coupling of electrons with lattice vibrations or spin fluctuations. Our findings indicate that doping-induced modifications to the Fermi surface and temperature-dependent renormalization play crucial roles in shaping the electronic structure. The results provide insights into the unconventional superconductivity of iron-based compounds, highlighting the importance of electron interactions in determining their unique properties.\\
 \midrule
 \textbf{Deepseek, 3-shot, ``We'' pronoun: 0.000000}\\\midrule
     The study investigates the electronic properties of hole-doped BaFe₂As₂, focusing on satellite features observed in spectroscopic measurements and their sensitivity to doping and temperature variations. Here's a structured summary: Material Context: Hole-Doping: Introduces excess positive charge carriers, altering the electronic structure significantly. Iron-Based Superconductor: Part of a class of materials known for high critical temperatures and complex electronic interactions. Key Observations: Satellite Features: These are secondary peaks near primary spectral lines, likely arising from many-body interactions such as electron-boson coupling. Doping Dependence: The intensity or prominence of these satellites varies with doping levels, indicating a correlation between carrier concentration and interaction strength. Temperature Dependence: Thermal effects modify the satellites, suggesting that phonon interactions or other thermally induced processes play a role. Implications for Superconductivity: Correlation with Superconducting Mechanisms: The presence and behavior of satellites may hint at the strength of pairing interactions crucial for superconductivity. Potential Optimization: Understanding how doping and temperature influence these features could guide strategies to enhance critical temperature (Tc). Methodology: Likely employs advanced spectroscopic techniques such as ARPES, supported by theoretical modeling to interpret the data accurately. Research Implications: Suggests a nuanced relationship between satellite features and superconductivity, highlighting the complexity of electronic interactions. Encourages further exploration across different superconducting materials to identify universal principles or unique mechanisms. This study contributes valuable insights into the intricate electronic dynamics of iron-based superconductors, offering potential pathways for advancing our understanding and engineering of high-Tc materials.\\
     \midrule
     \textbf{Deepseek, Style, ``We'' pronoun: 0.013605}\\\midrule
We report a comprehensive investigation into the electronic properties of hole-doped BaFe₂As₂, focusing on the significant influence of doping level and temperature variations. Our studies uncover a pronounced dopingand temperature-dependent behavior in the material's electronic characteristics, with particular emphasis on the emergence of satellite features in spectroscopic measurements. By employing advanced experimental techniques combined with theoretical modeling, we demonstrate how these satellite structures correlate with changes in carrier concentration and lattice dynamics. A detailed analysis reveals that the interplay between doping-induced charge carriers and thermal fluctuations plays a critical role in modulating the material's conductivity and band structure. This work provides fundamental insights into the underlying mechanisms governing the electronic response of doped iron-based compounds, offering valuable guidance for tailoring their properties for potential technological applications. Our findings underscore the importance of considering both doping and temperature effects when designing materials for optimized performance in electronic devices.\\
     \midrule
 \textbf{Deepseek, 0-shot CoT, ``We'' pronoun: 0.006993}\\\midrule
This study investigates the electronic properties of hole-doped BaFe₂As₂ using high-resolution angle-resolved photoemission spectroscopy (ARPES). We observe prominent satellite features in the spectral function, whose intensity and dispersion exhibit strong dopingand temperature-dependent behavior. These satellites, indicative of many-body interactions, become more pronounced with increased doping and lower temperatures, particularly near the Fermi level. Our analysis reveals a significant evolution of the electronic structure across the doping phase diagram, highlighting non-trivial renormalization effects and deviations from conventional quasiparticle behavior. The observed trends suggest that strong electron-electron correlations play a critical role in shaping the electronic properties of BaFe₂As₂. Furthermore, the dopingand temperature-dependent saturation of these satellites provides insights into the interplay between correlation effects and superconducting tendencies in iron-based superconductors. These findings challenge existing theoretical frameworks and underscore the importance of incorporating strong correlation physics in modeling the electronic states of doped iron pnictides.\\
\midrule
\textbf{Deepseek, 1-shot CoT, ``We'' pronoun: 0.007605}\\\midrule
In recent years, the study of strongly correlated electron systems has gained significant attention due to their unique electronic properties, particularly in materials exhibiting unconventional superconductivity. Hole-doped BaFe₂As₂, a member of the iron-based superconductor family, has emerged as a model system for investigating the interplay between electronic correlations, doping, and temperature-dependent phenomena. Understanding the intricate relationships between these factors is crucial for advancing our knowledge of quantum materials and their potential applications. Despite extensive research, the dopingand temperature-dependent evolution of electronic properties in BaFe₂As₂ remains poorly understood, particularly in the underdoped regime where strong correlations dominate. To address these challenges, we employ advanced spectroscopic techniques to probe the electronic structure of hole-doped BaFe₂As₂ across a wide range of doping levels and temperatures. Our measurements reveal pronounced satellite features in the electronic spectra, which are highly sensitive to both doping and temperature. These observations suggest the presence of strong many-body interactions, whose strength varies significantly with doping and thermal fluctuations. By systematically analyzing these dependencies, we uncover universal scaling behaviors that shed light on the underlying mechanisms governing the electronic response in this class of materials. Our findings demonstrate that the dopingand temperature-dependent evolution of electronic properties in BaFe₂As₂ cannot be explained by conventional theories alone, pointing to the importance of incorporating strong correlation effects. Furthermore, the emergence of robust satellite structures highlights the critical role of short-range Coulomb interactions in shaping the electronic landscape. These insights not only advance our understanding of the phase diagram of iron-based superconductors but also provide a framework for exploring similar phenomena in other strongly correlated electron systems.
\\
\midrule 
     \textbf{Deepseek, self-refine, ``We'' pronoun: 0.012658} \\\midrule
     Refined Hole-doped BaFe₂As₂ reveals fascinating insights into the interplay between strong electron correlations and unconventional superconductivity. Through advanced techniques like ARPES and transport measurements, we explored how doping and temperature affect its electronic properties. Notably, we identified significant satellite features in the valence band, which change markedly with doping and temperature, reflecting substantial particle interactions a hallmark of many-body physics. These changes align with shifts in the Fermi surface and quasiparticle behavior, underscoring how correlations redefine the material's electronic landscape. Moreover, our measurements showed variations in resistivity and Hall coefficients tied to doping and temperature, hinting at intricate interactions between charge carriers and spin fluctuations. Intriguingly, a nonlinear pattern emerged across the doping phase diagram, suggesting competing orders near quantum criticality. These findings highlight the pivotal role of strong correlations in iron pnictides and offer insights into exotic phases in similar materials. Such understanding could pave the way for innovative device technologies, bridging cutting-edge science with practical applications.\\
    \bottomrule
    \end{tabular}}
    \caption{Case study of human and AI texts for the cross-prompt generalization setting shown in Figure \ref{fig:cross-prompt-spec} (b).\\}\label{tab:case_study_b}
\end{table*}



\end{document}

%% file: tables/heatmap_corr_table.tex
\begin{table*}[ht]
\centering
\footnotesize
\begin{tabular}{lccccccc}
\toprule
& \multirow{2}{*}{\textbf{Feature Metric}}& \multicolumn{3}{c}{\textbf{DeBERTa-based}}&\multicolumn{3}{c}{\textbf{RoBERTa-based}}\\
\cmidrule(lr){3-5}\cmidrule(lr){6-8}
  &  & C-P &  C-M &  C-D &  C-P &  C-M &  C-D \\
\midrule
Readability & Gunning fog & 0.056 & \textbf{0.248} & \textbf{0.231} & 0.043 & \textbf{0.308} & \textit{\textbf{0.261}} \\
\midrule
Part-of-Speech & Numerals & \textit{\textbf{0.108}} & \textit{\textbf{0.363}} & 0.086 & \textbf{0.076} & \textbf{0.157} & 0.031 \\
\midrule
 & Passive voice & \textbf{\underline{\textbf{0.109}}} & \textbf{0.281} & \textbf{\underline{\textbf{0.296}}} & 0.054 & \textbf{0.221} & \textbf{\underline{\textbf{0.287}}} \\
 & Active voice & 0.010 & \textbf{0.194} & 0.066 & 0.061 & \textit{\textbf{0.373}} & 0.014 \\
Grammatical & Present tenses & 0.046 & \textbf{0.147} & 0.144 & \textbf{\underline{\textbf{0.116}}} & \textbf{0.328} & 0.173 \\
 & Past tenses & 0.018 & \textbf{\underline{\textbf{0.416}}} & 0.031 & 0.006 & \textbf{0.324} & 0.049 \\
 & Future tenses & \textbf{0.062} & 0.066 & 0.159 & \textbf{0.069} & \textbf{0.172} & 0.111 \\
\midrule
 & Personal names & \textbf{0.076} & \textit{\textbf{0.381}} & 0.030 & 0.046 & \textbf{0.182} & 0.071 \\
 & Adjectives in comparative degree & 0.027 & 0.123 & \textit{\textbf{0.267}} & 0.023 & \textbf{0.317} & \textit{\textbf{0.251}} \\
 & Adjectives in superlative degree & \textit{\textbf{0.095}} & \textbf{0.239} & 0.034 & 0.047 & 0.123 & 0.055 \\
Lexical & ``We'' pronoun & 0.014 & 0.041 & 0.015 & \textit{\textbf{0.108}} & 0.030 & 0.037 \\
 & 	``It'' pronoun & 0.029 & \textbf{0.236} & 0.184 & \textbf{0.077} & \textbf{\underline{\textbf{0.385}}} & 0.120 \\
 & ``Our'' possessive pronoun & 0.004 & 0.007 & \textit{\textbf{0.261}} & \textit{\textbf{0.113}} & 0.063 & \textbf{0.246} \\
 & ``Yourself'' pronoun & 0.030 & \textbf{0.183} & 0.157 & 0.061 & \textit{\textbf{0.366}} & 0.111 \\
\bottomrule
\end{tabular}
\caption{The \textbf{overall} Pearson correlation between generalization performance with different linguistic features. This table only presents the features that fall into the top 3 correlated features in one of the settings, more results are shown in Table \ref{tab:app-corr} (Appendix). The \textbf{significant (p<0.05) correlation is bolded}. We \underline{underline the strongest correlation} for each setting, and \textit{italicize the other scores within the top 3 correlated features}. The results of other features that are less correlated are shown in the Appendix. }
\label{tab:main-corr}
\end{table*}

%% file: tables/prompt_sep_corr_table.tex
\begin{table*}[htbp]
\centering
\footnotesize
\begin{tabular}{lccccccccccccccccc}
\toprule
&\multicolumn{5}{c}{\textbf{DeBERTa-based}}&\multicolumn{5}{c}{\textbf{RoBERTa-based}}\\
\cmidrule(lr){2-6}\cmidrule(lr){7-11}
 &  Abstracts & News &  Reviews &  QA &  ALL  & Abstracts & News &  Reviews &  QA &  ALL  \\
\midrule
Mistral 123B & \textbf{0.421} & \textbf{0.636} & \textbf{0.577} & \textbf{0.709} & \textbf{\underline{0.395}} & 0.155 & \textbf{\underline{0.703}} & \textbf{0.573} & 0\textbf{.630} & \textbf{0.219} \\
Deepseek 70B & \textbf{0.342} & \textbf{0.605} & \textbf{0.415} & \underline{\textbf{0.735}} & \textbf{0.276} & \textbf{0.528 }& \textbf{0.506} & \textbf{0.459} & \textbf{\underline{0.728}} & \textbf{0.400} \\
Llama 70B & \textbf{0.412} & 0.209 & \underline{\textbf{0.736}} & \textbf{0.584} & \textbf{0.346} & 0.245 & 0.248 & \textbf{\underline{0.758}} & \textbf{0.572} & \textbf{0.380} \\
Qwen 72B & 0.072 & \textbf{0.590} & 0.317 & 0.301 & 0.098 & 0.128 & \textbf{0.519} & \textbf{0.549} & \textbf{0.357} & \textbf{0.212} \\
Qwen 32B & 0.214 & \underline{\textbf{0.684}} & \textbf{0.527} & \textbf{0.678} & \textbf{0.183} & \textbf{0.377} & \textbf{0.672} & \textbf{0.617} & \textbf{0.492} & \textbf{0.308} \\
Qwen 14B & 0.275 & \textbf{0.476} & \textbf{0.553} & \textbf{0.560 }& \textbf{0.218} & 0.264 & \textbf{0.457} & \textbf{0.605} & \textbf{0.580} & \textbf{0.204} \\
Solar 22B & \underline{\textbf{0.563}} & \textbf{0.482} & \textbf{0.517} & \textbf{0.614} & \textbf{0.303} & \textbf{\underline{0.703}} & \textbf{0.688} & 0.320 & \textbf{0.582} & \textbf{\underline{0.426}} \\
ALL & \textbf{0.196} & \textbf{0.231} & \textbf{0.437} & \textbf{0.255} & \textbf{0.109} & \textbf{0.246} & \textbf{0.155} & \textbf{0.486} & \textbf{0.224} & \textbf{0.116} \\
\bottomrule
\end{tabular}
\caption{The cross-prompt correlation between generation performance and the most correlated linguistic feature when evaluated on different datasets and models. The \textbf{significant correlation is bolded}. We \underline{underline the strongest correlation} for each dataset. Similar cross-model and cross-dataset results are in Appendix.}
\label{tab:cross-prompt-sep-corr}
\end{table*}

%% file: prompt_detail.tex
\subsection{Overview of prompting techniques} \label{prompt-appendix}
To test diverse prompting techniques, we chose six different prompt types. The prompt types are:

\textbf{0-shot}: Our zero-shot prompts are simple instructions for the language models that include a basic piece of information about the text and desired length of the text. For example, the template for the prompt to generate abstracts is \textit{Write an abstract for an article titled “\{title\}”. The abstract should be around \{length\} characters long.}

\textbf{Chain-of-Thought (0-shot CoT)}:  Chain-of-thought prompting is a strategy that aims to improve reasoning by dividing the task into smaller steps \citep{wei2022chain}. A chain-of-thought prompt usually relies on exemplars, containing a prompt and a correct response. The example response is expressed as a series of steps that lead to a final output. This prompts the model to reason step-by-step. 

For the purpose of our dataset, we adapt a \textbf{1-shot CoT} prompting strategy to the task of text generation: this prompt subtype consists of a human-written step-by-step instruction for the model. The instruction is based on an example of a human-written text from the dataset. 

We also use \textbf{0-shot CoT} prompts \citep{kojima2023largelanguagemodelszeroshot}, which consist of adding \textit{let's think step by step} to the baseline prompt to cause step-by-step reasoning.

\textbf{Few-shot In-context learning (3-shot)} \citep{brown2020languagemodelsfewshotlearners}: The few-shot prompt contains several examples of input-output pairs for a given task. Our few-shot prompts include three human-written examples from the dataset.

\textbf{Style examples (Style)}: In \citet{10.1007/978-3-031-57850-2_16}, prompts with style guidelines have been effective in prompting LLMs to generate output that evades detection methods, and models have been prompted to simulate the styles of several famous writers. We adapt one of the prompts from \citet{10.1007/978-3-031-57850-2_16} for our task. Specifically, we use the style example prompt, but with an example from the human-written part of our dataset. Even though using a text written by a famous author (e.g., Shakespeare) as a style example has been successful in preventing detection \cite{10.1007/978-3-031-57850-2_16}, our prompt modification aims to create a more realistic setting to adapt to different datasets.

\textbf{Self-refine} \citep{madaan2023selfrefineiterativerefinementselffeedback}: Self-refinement prompts are prompts that use the LLM itself to critique and improve its own responses. The prompts of this kind are comprised of three stages: generating the output, generating feedback for the output and applying the feedback to the output. The second and third steps are repeated until a stopping condition is met. In this case, we base the stopping condition on the ability of the model itself to distinguish its own outputs from human-written text. To achieve this, we use evaluation prompts based on the GPT-4 evaluation prompt used in \citep{madaan2023selfrefineiterativerefinementselffeedback}. For the purpose of our task, we ask the model to decide which text sounds more human-written.
    The resulting set of prompts is made of the following stages:
\begin{itemize}
    \item Initialization prompt: the same as our baseline zero-shot prompt.
    \item Feedback prompt: prompt used for generating feedback on how to make the text seem more human-written.
    \item Iterate prompt: used to get the next iteration if the model classifies its generated text as AI.
    \item Evaluate prompt: used to check whether the stopping condition (the model classifies the text as human-written) has been met.
\end{itemize}

%% file: linguistic_analysis_appendix.tex
\paragraph{Lexical diversity.}
We choose Moving Average Type-Token Ratio to measure the lexical diversity of the texts, as it is independent of the text length \citep{Covington01052010}. Additionally, we compare the texts in terms of the \textbf{number of unique words}, as this feature has been shown to be relevant in the previous research \cite{opara2024styloaidistinguishingaigeneratedcontent, yildiz_durak_comparison_2025}.

\paragraph{Lexical density.} Lexical density is the percentage of content words in the text. Machine-generated text is said to achieve higher lexical density \citep{savoy_machine_2020} than human-written text, which means that it contains fewer function words. The content words are adjectives, adverbs, nouns, and verbs. 

\paragraph{Sentiment.} AI-generated text has been shown to differ from human-written text in terms of sentiment: some authors have written of \textit{positivity bias} present in AI-generated texts \cite{doi:10.1177/0261927X231200201, munoz-ortiz_contrasting_2023, margolina_exploring_2023}, which means that text generated by large language models tends to contain more positive emotions compared to human-written texts. Other research suggests that human-written texts are more varied in terms of the richness of emotional content \cite{zanotto2024humanvariabilityvsmachine}. We use the TextBlob library to calculate the \textbf{polarity} and \textbf{subjectivity} scores for the texts in our dataset. \textbf{Polarity} is a score between -1 and 1, where -1 denotes a negative sentiment, while 1 denotes a positive sentiment. \textbf{Subjectivity} relates to the amount of personal opinion included in the text.

\paragraph{Readability.} Readability refers to the ease of understanding a text. AI-generated text tends to be less readable than human-written text \cite{yadagiri-etal-2024-detecting,doi:10.1177/0261927X231200201,10545131}. We use the \textbf{Gunning fog index} and the \textbf{Flesch reading ease test} as measures of readability. The \textbf{Gunning fog index} is an estimated number of years needed to understand a given passage. The \textbf{Flesch reading ease test} is a metric of readability, in which texts that are easier to read receive a higher score. In the analysis of readability, we also include the \textbf{text length} in characters, as well as sentence length statistics. Machine-generated text tends to be less varied than human-written text in terms of sentence length \citep{DESAIRE2023101426,munoz-ortiz_contrasting_2023}. Following \citep{DESAIRE2023101426}, we calculate the \textbf{average sentence length}, the \textbf{standard deviation from the average sentence length}, and the \textbf{number of very long} (35 words or more) and \textbf{very short} (10 words or less) \textbf{sentences} in each text.

\paragraph{Part-of-Speech (POS).} We use a selection of metrics from StyloMetrix \cite{okulska2023stylometrixopensourcemultilingualtool} to compare the frequency of verbs, nouns, adjectives, pronouns, determiners, conjunctions and numerals across the different dataset types, models and prompts. The frequency of parts of speech is measured as the fraction of text covered by tokens representing a given part of speech. The frequency of POS has been shown to be different across different text genres, with non-fiction texts typically achieving a higher frequency of nouns than fiction texts \cite{MendhakarHS+2023+99+131}. A high frequency of verbs can be associated with more narrative texts, while a high frequency of adjectives is common for more descriptive texts. Additionally, previous research has discovered differences between human-written and AI-generated text in terms of frequency of certain POS \cite{georgiou2024differentiatinghumanwrittenaigeneratedtexts}. Therefore, POS analysis is relevant for gaining insights into the literary style of texts in our dataset.

\begin{figure*}[ht]
    \centering
    \includegraphics[width=1\linewidth]{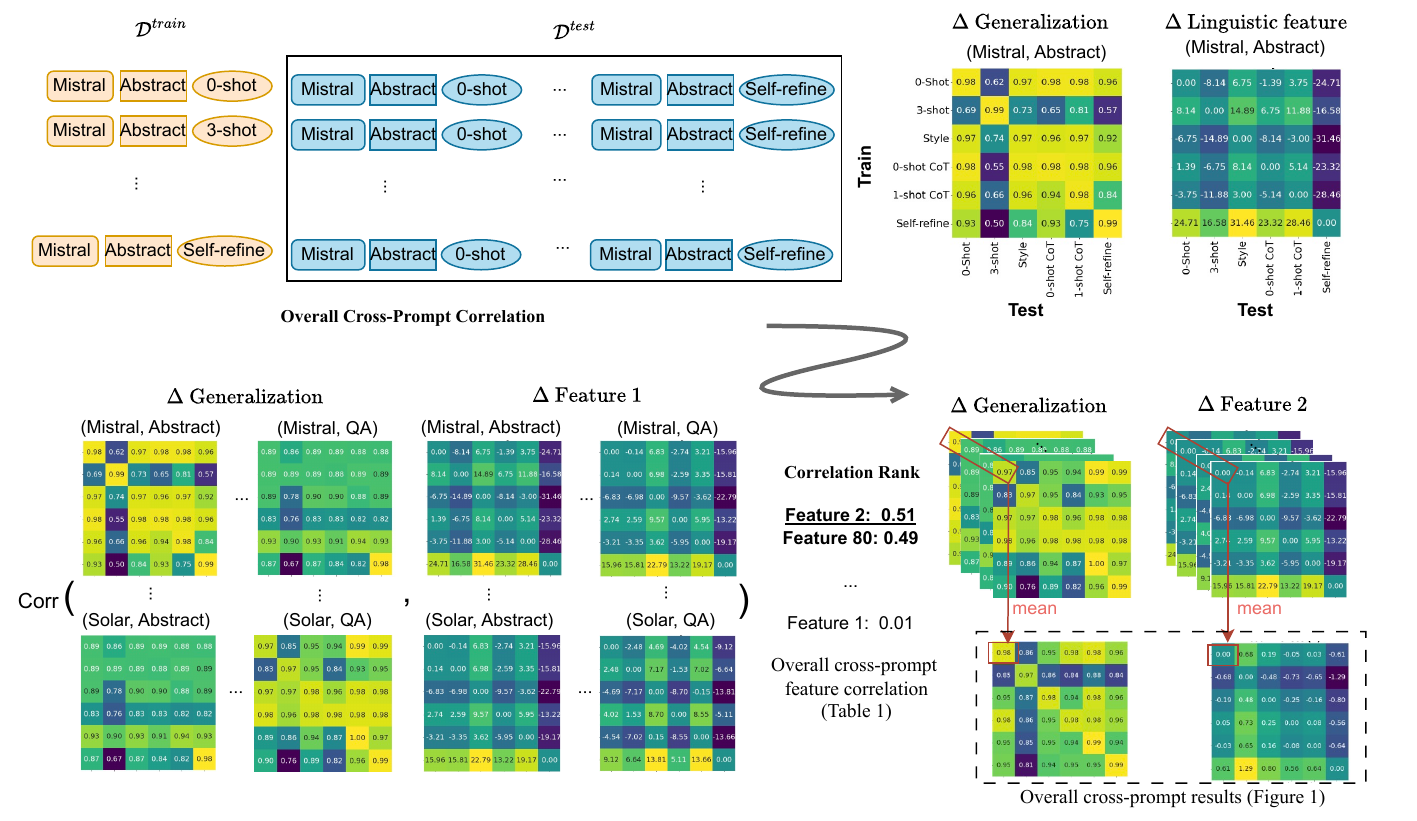}
    \caption{The workflow to get the overall cross-prompt generalization and feature shift results.}
    \label{fig:enter-label}
\end{figure*}
\input{tables/cross-model-sep-table}
\input{tables/cross-data-sep-table}
\paragraph{Grammatical analysis.} We use StyloMetrix \cite{okulska2023stylometrixopensourcemultilingualtool} metrics to compare texts in terms of grammatical categories related to verbs. In previous research, human-written texts have been shown to contain more passive voice than AI-generated texts \cite{georgiou2024differentiatinghumanwrittenaigeneratedtexts}. We measure the \textbf{incidence of passive and active voice} as the frequency of verbs in passive or active voice. We also compare the differences between the choice of tenses. The \textbf{frequency of past, present and future tenses} is measured as the fraction of the text covered by verbs in past, present and future tenses. For example, the incidence of past tenses is the number of verbs in past simple, past continuous, past perfect or past perfect continuous divided by the total number of words in the text.


\paragraph{Lexical analysis.} As part of lexical analysis, we measure the \textbf{frequency of proper and personal names}, as well as the \textbf{frequency of adjectives and adverbs in positive, comparative and superlative degrees} and the \textbf{frequency of nouns in possessive case}.
We use the StyloMetrix \cite{okulska2023stylometrixopensourcemultilingualtool} pronoun-related metrics to analyze the differences in the usage of pronouns in human-written and AI-generated text, as well as across the different prompt types, dataset types and models. We calculate not only the frequency of pronouns (POS\_PRO), but also the frequency of specific personal or reflexive pronouns and the general frequency of certain types of pronouns (for example, the frequency of all first-person singular pronouns).

%% file: tables/cross-model-sep-table.tex
\begin{table*}[hbpt]
\centering
\footnotesize
\begin{tabular}{lcccccccccccccccccccc}
\toprule

&\multicolumn{5}{c}{DeBERTa-based}&\multicolumn{5}{c}{RoBERTa-based}\\
 &  Abstracts & News &  Reviews &  QA &  ALL  & Abstracts & News &  Reviews &  QA &  ALL  \\
\midrule
0-shot &{0.587} & {0.569} & {0.663} & {0.591}  &{0.416}  & {0.399} & {0.677} & {0.655} & {0.355} & {0.385}\\
\bottomrule
\end{tabular}
\caption{The cross-model correlation between generation performance and the most correlated linguis-
tic feature when evaluated on different datasets.}\label{tab:cross-model-sep-corr}
\end{table*}

%% file: tables/cross-data-sep-table.tex
\begin{table}[ht]
\centering
\footnotesize
\begin{tabular}{lcccccccccccccccccccc}
\toprule
& DeBERTa-based & RoBERTa-based \\
& 0-shot & 0-shot \\
\midrule
Mistral 123B&0.310 & 0.350 \\
Deepseek 70B&0.418 & 0.468 \\
Llama 70B&0.389 & 0.374 \\
Qwen 72B&0.425 & 0.308 \\
Qwen 32B&0.426 & 0.418 \\
Qwen 14B&0.487 & 0.545 \\
Solar 22B&0.269 & 0.188 \\
ALL &0.296 & 0.287 \\
\bottomrule
\end{tabular}
\caption{The cross-dataset correlation between generation performance and the most correlated linguis-
tic feature when evaluated on different models. }
\label{tab:cross-dataset-sep-corr}
\end{table}

%% file: tables/prompt_templates.tex
\onecolumn
\scriptsize
\begin{longtable}{|l|l|p{12.5cm}|} 
\toprule
\textbf{Prompt } & \textbf{Dataset} & \textbf{Prompt template} \\ \midrule
\endfirsthead


\toprule
\textbf{Prompt } & \textbf{Dataset} & \textbf{Prompt template} \\
\midrule
\endhead

\midrule
\multicolumn{3}{r}{\small Continued on next page} \\
\endfoot

\bottomrule
\addlinespace[10pt]
\caption{Prompt details.}\label{tab:promp-detail} \\
\endlastfoot
  
\multirow{4}{*}{0-shot} &
  Abstracts &
  "Write an abstract for an article in \{category\} with a title: \textbackslash{}"\{title\}\textbackslash{}". The abstract should be around \{length\} characters long." \\ \cline{2-3} 
 &
  News &
  "Write a news article based on the following highlights:\textbackslash{}n\textbackslash{}"\{highlights\}\textbackslash{}"\textbackslash{}nYour article should be around \{length\} characters long." \\ \cline{2-3} 
 &
  Reviews &
  "Write an Amazon review for the item \textbackslash{}"\{item\_name\}\textbackslash{}" with a title \textbackslash{}"\{title\}\textbackslash{}" and a rating of \{rating\}. The review should be around \{length\} characters long." \\ \cline{2-3} 
 &
  QA &
  "\{question\}\textbackslash{}n Your answer should be around \{length\} characters long." \\ \hline
\multirow{4}{*}{0-shot CoT} &
  Abstracts &
  "Write an abstract for an article in \{category\} with a title: \textbackslash{}"\{title\}\textbackslash{}". Let’s think step by step. Your answer should only include the abstract. The abstract should be around \{length\} characters long." \\ \cline{2-3} 
 &
  News &
  "Write a news article based on the following highlights:\textbackslash{}n\textbackslash{}"\{highlights\}\textbackslash{}"\textbackslash{}nLet's think step by step. Your answer should only include the article. The article should be around \{length\} characters long." \\ \cline{2-3} 
 &
  Reviews &
  "Write an Amazon review for the item \textbackslash{}"\{item\_name\}\textbackslash{}" with a title \textbackslash{}"\{title\}\textbackslash{}" and a rating of \{rating\}. Let’s think step by step. Your answer should only include the review. The review should be around \{length\} characters long." \\ \cline{2-3} 
 &
  QA &
  "\{question\}\textbackslash{}nLet’s think step by step. Your answer should only include the answer to the question. The answer should be around \{length\} characters long." \\ \hline

\multirow{4}{*}{1-shot CoT} &
  Abstracts &
  "I want to write an abstract for an article in computer science. The article is titled \textbackslash{}"ConQRet: Benchmarking Fine-Grained Evaluation of Retrieval Augmented Argumentation with LLM Judges\textbackslash{}". \textbackslash{}n 1. First, I introduce the context of the research and explain the motivation:\textbackslash{}n\textbackslash{}"Computational argumentation, which involves generating answers or summaries for controversial topics like abortion bans and vaccination, has become increasingly important in today's polarized environment. Sophisticated LLM capabilities offer the potential to provide nuanced, evidence-based answers to such questions through Retrieval-Augmented Argumentation (RAArg), leveraging real-world evidence for high-quality, grounded arguments. However, evaluating RAArg remains challenging, as human evaluation is costly and difficult for complex, lengthy answers on complicated topics. At the same time, re-using existing argumentation datasets is no longer sufficient, as they lack long, complex arguments and realistic evidence from potentially misleading sources, limiting holistic evaluation of retrieval effectiveness and argument quality.\textbackslash{}"\textbackslash{}nThen, I describe how I addressed the gaps in current research and give a detailed description of my methodology:\textbackslash{}n\textbackslash{}"To address these gaps, we investigate automated evaluation methods using multiple fine-grained LLM judges, providing better and more interpretable assessments than traditional single-score metrics and even previously reported human crowdsourcing. To validate the proposed techniques, we introduce ConQRet, a new benchmark featuring long and complex human-authored arguments on debated topics, grounded in real-world websites, allowing an exhaustive evaluation across retrieval effectiveness, argument quality, and groundedness. We validate our LLM Judges on a prior dataset and the new ConQRet benchmark.\textbackslash{}"\textbackslash{}nFinally, I describe the results and their implications for the research on this topic:\textbackslash{}n\textbackslash{}"Our proposed LLM Judges and the ConQRet benchmark can enable rapid progress in computational argumentation and can be naturally extended to other complex retrieval-augmented generation tasks.\textbackslash{}"\textbackslash{}nBased on the provided step-by-step instruction, write an abstract for an article in \{category\} titled \textbackslash{}"\{title\}\textbackslash{}"." \\ \cline{2-3} 

 &
  News &
  "I want to write a news article about the following events:\textbackslash{}nDarsh Patel, 22, was hiking with friends in the Apshawa Preserve in West Milford on Sunday when a bear started following them.\textbackslash{}nThe group fled in different directions and when the four other hikers could not find Patel, they called police.\textbackslash{}nPatel's body was found two hours later.\textbackslash{}nThe 300-pound bear was circling the body and could not be scared away.\textbackslash{}nIt was shot dead in accordance with Division of Fish and Wildlife guidelines.\textbackslash{}nOn Saturday locals splitting wood filmed a bear rifling through their garbage.\textbackslash{}n\textbackslash{}n\textbackslash{}nFirst, I introduce the event and its content to the readers:\textbackslash{}nLocals in northern New Jersey believe they filmed a black bear hunting for food hours before a 22-year-old hiker was mauled to death in nearby woods at the weekend. Two men splitting wood on Saturday captured a video of a bear going through garbage just a few feet from where they were working, before scampering off into the woods, according to CNN. On Sunday, Darsh Patel, a senior majoring in information technology and informatics at Rutgers University, was found dead in Apshawa Preserve - about 45 miles northwest of New York City - with a 300-pound bear guarding his body. Officials say the attack was the first fatal bear-human encounter on record in New Jersey. Just a day after the footage was shot, a black bear mauled a 22-year-old student to death in the woods nearby.\textbackslash{}nThen, I provide more details related to the event:\textbackslash{}nPatel had been hiking with four friends in the 526-acre woods. The five friends noticed the bear beginning to follow them and ran, splitting up as they did. When they couldn't find Patel, they called police, who found his body about two hours later. The bear was about 30 yards from the body and circling, Department of Environmental Protection spokesman Larry Ragonese said, and wouldn't leave even after officers tried to scare it away by making loud noises and throwing sticks and stones. The male bear was killed with two rifle blasts and is being examined at a state lab for more clues as to why it may have pursued the group of five hikers.\textbackslash{}nThen, I provide opinions on the event, quoting officials, experts or witnesses of the event:\textbackslash{}nKelcey Burguess, principal biologist and leader of the state Division of Fish and Wildlife's black bear project, said the bear could have been predisposed to attack but more likely was looking for food. State and local officials stressed that bear attacks are rare even in a region of the state that may have as many as 2,400 bruins in its dense forests. \textbackslash{}"This is a rare occurrence,\textbackslash{}" West Milford police Chief Timothy Storbeck said, noting that his department receives six to 12 calls per week regarding bears, usually involving them breaking into trash cans. Locals: Residents in northern New Jersey often spot bears in and around their yards. There are as many 2,400 bruins in the area's dense forests, but until now had never been a fatal human-bear attack. Wildlife officials believe there is a current shortage of the acorns and berries that bears eat. The hikers had granola bars and water with them, Storbeck said. Officials don't believe the hikers provoked the bear but they may have showed their inexperience when they decided to run. The safest way to handle a bear encounter is to move slowly and not look the bear in the eye, DEP spokesman Larry Ragonese said. New Jersey Division of Fish and Wildlife guidelines direct law enforcement to euthanize \textbackslash{}"Category I\textbackslash{}" bears, which are deemed an \textbackslash{}"immediate threat to human safety.\textbackslash{}" NJ Advance Media reports that the New Jersey State Medical Examiner, the Fish and Wildlife Division of the state Department of Environmental Protection and the West Milford Police Department are looking into the circumstances of Patel's death.\textbackslash{}nFinally, I conclude the report by highlighting the relevance of the event:\textbackslash{}n\textbackslash{}"Bear sightings are not unusual by any stretch in New Jersey,\textbackslash{}" said Bob Considine, spokesperson for the Department of Environmental Protection. \textbackslash{}"They have been seen in all 21 counties, although they’re obviously most common in the northwest part of the state.\textbackslash{}" Black bears rarely pose a threat to humans and often retreat when confronted. In 2006, a tabby cat scared a black bear up a tree in West Milford. The bear only climbed down and left after the cat's owner had called it back into the house.\textbackslash{}n\textbackslash{}nNow, I want to write an article on a different topic:\textbackslash{}n\{highlights\}\textbackslash{}nWrite the article, following the steps described above. Your answer should only include the article. The article should be around \{length\} characters long.", \\ \cline{2-3} 
 &
  Reviews &
  "I want to write an Amazon review for an item called \textbackslash{}"Haier RDG350AW 6.5 Cubic Foot Front Load Gas Dryer, White\textbackslash{}". The rating will be 5.0.\textbackslash{}nFirst, I choose a title for my review that describes my opinion and experience well. The title of my review will be:\textbackslash{}n\textbackslash{}"Very Affordable Dryer\textbackslash{}".\textbackslash{}nThen, I would state my initial experience with the product:\textbackslash{}n\textbackslash{}"I was a little worried about buying this cause it had some bad reviews, but it's a really great deal.\textbackslash{}"\textbackslash{}nThen I would describe the pros and cons of the product, expressing the reasons for my rating:\textbackslash{}n\textbackslash{}"It didn't touch my gas bill period. Yes larger loads take a while to dry, maybe up to 3 to 4 hours. It's just really energy efficent. Like I said my gas bill didn't budge with this being hooked up.\textbackslash{}"\textbackslash{}nNow, I want to write a review of an item called \textbackslash{}"\{item\_name\}\textbackslash{}" and give it a rating of \{rating\}.\textbackslash{}nFollow the directions described above to come up with the review. Your answer should only include the review and the title. The review should be one paragraph." \\ \cline{2-3} 
 &
  QA &
  "If I was asked the question: \textbackslash{}"When do the English state schools finish summer term and holiday begins?\textbackslash{}",\textbackslash{}nFirst, I would give a general overview of the answer:\textbackslash{}"In the English school system, state schools run from early September to mid or late July of the following year.\textbackslash{}"\textbackslash{}nThen, I would go into more detail:\textbackslash{}n\textbackslash{}"The summer term (also known as the third term) runs from late April and finishes mid to late July with a week-long half term break in between. The summer holiday begins in late July and usually runs about six weeks long, ending in September.\textbackslash{}"\textbackslash{}nFinally, I would add nuance or additional context to my answer:\textbackslash{}n\textbackslash{}"The schools on the Trinity terms end their school year and begin summer holidays a few weeks earlier, at the end of June.\textbackslash{}"\textbackslash{}nFollowing the steps described above, answer the question: \textbackslash{}"\{question\}\textbackslash{}" in one paragraph." \\ \hline

\multirow{4}{*}{3-shot} &
  Abstracts &
  \begin{tabular}[c]{@{}l@{}}"role" : "user", "content" : \{title\_1\}\\ "role" : "assistant", "content" : \{human-written\_abstract\_1\}\\ "role" : "user", "content" : \{title\_2\}\\ "role" : "assistant", "content" : \{human-written\_abstract\_2\}\\ "role" : "user", "content" : \{title\_3\}\\ "role" : "assistant", "content" : \{human-written\_abstract\_3\}\\ "role" : "user", "content" : \{title\}\end{tabular} \\ \cline{2-3} 
 &
  News &
  \begin{tabular}[c]{@{}l@{}}"role" : "user", "content" : \{highlights\_1\}\\ "role" : "assistant", "content" : \{human-written\_article\_1\}\\ "role" : "user", "content" : \{highlights\_2\}\\ "role" : "assistant", "content" : \{human-written\_article\_2\}\\ "role" : "user", "content" : \{highlights\_3\}\\ "role" : "assistant", "content" : \{human-written\_article\_3\}\\ "role" : "user", "content" : \{highlights\}\end{tabular} \\ \cline{2-3} 
 &
  Reviews &
  \begin{tabular}[c]{@{}l@{}}"role" : "user", "content" : \{product\_name\_1\}\\ "role" : "assistant", "content" : \{human-written\_review\_1\}\\ "role" : "user", "content" : \{product\_name\_2\}\\ "role" : "assistant", "content" : \{human-written\_review\_2\}\\ "role" : "user", "content" : \{product\_3\}\\ "role" : "assistant", "content" : \{human-written\_review\_3\}\\ "role" : "user", "content" : \{product\}\end{tabular} \\ \cline{2-3} 
 &
  QA &
  \begin{tabular}[c]{@{}l@{}}"role" : "user", "content" : \{question\_1\}\\ "role" : "assistant", "content" : \{human-written\_answer\_1\}\\ "role" : "user", "content" : \{question\_2\}\\ "role" : "assistant", "content" : \{human-written\_answer\_2\}\\ "role" : "user", "content" : \{question\_3\}\\ "role" : "assistant", "content" : \{human-written\_answer\_3\}\\ "role" : "user", "content" : \{question\}\end{tabular} \\ \hline
\multirow{4}{*}{Style} &
  Abstracts &
  "As an academic paper writer, your task is to write an abstract of a research paper in a specific writing style. Write in the writing style of an example but ignore the content and topic of the example. You will be provided with the style example. You will be provided with the title for your abstract.\textbackslash{}nStyle example: \{example\}\textbackslash{}nTitle: \{title\}" \\ \cline{2-3} 
 &
  News &
  "As a news article writer, your task is to write a news article in a specific writing style. Write in the writing style of an example but ignore the content and topic of the example. You will be provided with the style example. You will be provided with the summary of the topic for your article.\textbackslash{}nStyle example: \{example\}\textbackslash{}nSummary: \{highlights\}" \\ \cline{2-3} 
 &
  Reviews &
  "As an Amazon review writer, your task is to write a review for an item in a specific writing style. Write in the writing style of an example but ignore the content and topic of the example. You will be provided with the style example. You will be provided with the name of the item you have to review.\textbackslash{}nStyle example: \{example\}\textbackslash{}nItem: \{item\_name\}" \\ \cline{2-3} 
 &
  QA &
  "As a highly intelligent question answering bot, your task is to answer questions in specific writing styles. Write in the writing style of an example but ignore the content and topic of the example. You will be provided with the style example. You will be provided with the question.\textbackslash{}nStyle example: \{example\}\textbackslash{}nQuestion: \{question\}" \\ \hline
\multirow{4}{*}{Self-refine} &
  Abstracts &
  \begin{enumerate}
      \item "Write an abstract for an article in \{category\} with a title: \textbackslash{}"\{title\}\textbackslash{}". The abstract should be around \{length\} characters long."
      \item "You will see an abstract for a scientific article. Your task is to provide feedback on how to make the text seem more human-like. Consider sentence length, sentence structure, vocabulary and readability.\textbackslash{}nAbstract: \{text\}\textbackslash{}nFeedback: "
      \item "Based on the feedback, improve the text below:\textbackslash{}nText: \{text\}\textbackslash{}nFeedback: \{feedback\}."
      \item "Which text sounds more human-written?\textbackslash{}nText A: \{text\_a\}\textbackslash{}nText B: \{text\_b\}\textbackslash{}n\textbackslash{}nPick your answer from {[}\textbackslash{}"Text A\textbackslash{}", \textbackslash{}"Text B\textbackslash{}", \textbackslash{}"both\textbackslash{}", \textbackslash{}"neither\textbackslash{}"{]}. Generate a short explanation for your choice first. Then, generate \textbackslash{}"Text A seems more human-written\textbackslash{}" or \textbackslash{}"Text B seems more human-written\textbackslash{}" or \textbackslash{}"Both texts seem human-written\textbackslash{}" or \textbackslash{}"Neither of the texts sounds human-written\textbackslash{}"
    \end{enumerate} \\ \cline{2-3} 
 &
  News &\begin{enumerate}
    \item  "Write a news article based on the following highlights:\textbackslash{}n\textbackslash{}"\{highlights\}\textbackslash{}"\textbackslash{}nYour article should be around \{length\} characters long."
    \item "You will see a news article. Your task is to provide feedback on how to make the text seem more human-like. Consider sentence length, sentence structure, vocabulary and readability.\textbackslash{}nArticle: \{text\}\textbackslash{}nFeedback: "
    \item "Based on the feedback, improve the text below:\textbackslash{}nText: \{text\}\textbackslash{}nFeedback: \{feedback\}."
    \item "Which text sounds more human-written?\textbackslash{}nText A: \{text\_a\}\textbackslash{}nText B: \{text\_b\}\textbackslash{}n\textbackslash{}nPick your answer from {[}\textbackslash{}"Text A\textbackslash{}", \textbackslash{}"Text B\textbackslash{}", \textbackslash{}"both\textbackslash{}", \textbackslash{}"neither\textbackslash{}"{]}. Generate a short explanation for your choice first. Then, generate \textbackslash{}"Text A seems more human-written\textbackslash{}" or \textbackslash{}"Text B seems more human-written\textbackslash{}" or \textbackslash{}"Both texts seem human-written\textbackslash{}" or \textbackslash{}"Neither of the texts sounds human-written\textbackslash{}"
\end{enumerate} \\ \cline{2-3} 
 &
  Reviews &
  \begin{enumerate}
      \item "Write an Amazon review for the item \textbackslash{}"\{item\_name\}\textbackslash{}" with a title \textbackslash{}"\{title\}\textbackslash{}" and a rating of \{rating\}. The review should be around \{length\} characters long."
      \item "You will see an Amazon review. Your task is to provide feedback on how to make the text seem more human-like. Consider sentence length, sentence structure, vocabulary and readability.\textbackslash{}nReview: \{text\}\textbackslash{}nFeedback: "
      \item "Based on the feedback, improve the text below:\textbackslash{}nText: \{text\}\textbackslash{}nFeedback: \{feedback\}."
      \item "Which text sounds more human-written?\textbackslash{}nText A: \{text\_a\}\textbackslash{}nText B: \{text\_b\}\textbackslash{}n\textbackslash{}nPick your answer from {[}\textbackslash{}"Text A\textbackslash{}", \textbackslash{}"Text B\textbackslash{}", \textbackslash{}"both\textbackslash{}", \textbackslash{}"neither\textbackslash{}"{]}. Generate a short explanation for your choice first. Then, generate \textbackslash{}"Text A seems more human-written\textbackslash{}" or \textbackslash{}"Text B seems more human-written\textbackslash{}" or \textbackslash{}"Both texts seem human-written\textbackslash{}" or \textbackslash{}"Neither of the texts sounds human-written\textbackslash{}"
\end{enumerate} \\ \cline{2-3} 
 &
  QA &
  \begin{enumerate}
      \item "\{question\}\textbackslash{}n Your answer should be around \{length\} characters long."
      \item "You will see an answer to a question. Your task is to provide feedback on how to make the text seem more human-like. Consider sentence length, sentence structure, vocabulary and readability.\textbackslash{}nAnswer: \{text\}\textbackslash{}nFeedback: "
      \item "Based on the feedback, improve the text below:\textbackslash{}nText: \{text\}\textbackslash{}nFeedback: \{feedback\}."
      \item "Which text sounds more human-written?\textbackslash{}nText A: \{text\_a\}\textbackslash{}nText B: \{text\_b\}\textbackslash{}n\textbackslash{}nPick your answer from {[}\textbackslash{}"Text A\textbackslash{}", \textbackslash{}"Text B\textbackslash{}", \textbackslash{}"both\textbackslash{}", \textbackslash{}"neither\textbackslash{}"{]}. Generate a short explanation for your choice first. Then, generate \textbackslash{}"Text A seems more human-written\textbackslash{}" or \textbackslash{}"Text B seems more human-written\textbackslash{}" or \textbackslash{}"Both texts seem human-written\textbackslash{}" or \textbackslash{}"Neither of the texts sounds human-written\textbackslash{}"
      \end{enumerate} 
\end{longtable}
\twocolumn

%% file: tables/main_results_table.tex
\begin{table*}[]
\adjustbox{max width=\textwidth}{%
\centering
\footnotesize
\begin{tabular}{lllccccccc}
\toprule
Detector & Model & Dataset & \multicolumn{6}{c}{Prompt type} \\
\cmidrule(lr){4-9}
 &  &  & 0-Shot & 3-Shot CoT & 1-Shot CoT & Style & 3-Shot & Self-refine \\
\midrule
\multirow{28}{*}{DeBERTa} & \multirow{4}{*}{Llama3.3 70b} & Abstracts & 0.988 & 0.9875 & 0.993 & 0.9795 & 0.9945 & 0.9865 \\
 &  & News & 0.999 & 0.999 & 0.9985 & 0.9775 & 0.7935 & 1.0 \\
 &  & Reviews & 0.984 & 0.988 & 0.9995 & 0.9615 & 0.9805 & 0.9875 \\
 &  & QA & 0.8866 & 0.8966 & 0.9989 & 0.9604 & 0.8233 & 0.9509 \\
\cmidrule(lr){2-9}
 & \multirow{4}{*}{Qwen 14b} & Abstracts & 0.998 & 0.991 & 0.994 & 0.998 & 0.999 & 0.999 \\
 &  & News & 1.0 & 0.9995 & 0.9995 & 1.0 & 1.0 & 0.999 \\
 &  & Reviews & 0.9985 & 0.996 & 1.0 & 1.0 & 0.9965 & 0.9985 \\
 &  & QA & 0.98 & 0.9662 & 0.9916 & 0.9926 & 0.9668 & 0.9736 \\
\cmidrule(lr){2-9}
 & \multirow{4}{*}{Qwen 32b} & Abstracts & 0.9955 & 0.9915 & 0.994 & 0.9965 & 1.0 & 0.999 \\
 &  & News & 1.0 & 0.999 & 0.9995 & 1.0 & 1.0 & 1.0 \\
 &  & Reviews & 0.999 & 0.9955 & 1.0 & 0.998 & 0.9985 & 0.9955 \\
 &  & QA & 0.9763 & 0.981 & 0.9852 & 0.9942 & 0.9626 & 0.9736 \\
\cmidrule(lr){2-9}
 & \multirow{4}{*}{Qwen 72b} & Abstracts & 0.988 & 0.9845 & 0.9845 & 0.991 & 0.9955 & 0.991 \\
 &  & News & 0.999 & 0.9995 & 0.9995 & 1.0 & 0.998 & 0.999 \\
 &  & Reviews & 0.995 & 0.9975 & 0.9995 & 0.9945 & 0.9965 & 0.989 \\
 &  & QA & 0.9699 & 0.9583 & 0.9889 & 0.942 & 0.8903 & 0.9662 \\
\cmidrule(lr){2-9}
 & \multirow{4}{*}{Solar 22b} & Abstracts & 0.9855 & 0.9915 & 0.992 & 0.989 & 0.9985 & 0.9975 \\
 &  & News & 1.0 & 0.9995 & 0.999 & 1.0 & 0.9994 & 0.9995 \\
 &  & Reviews & 0.998 & 0.998 & 0.9995 & 0.9975 & 0.9965 & 0.999 \\
 &  & QA & 0.9626 & 0.9852 & 0.9947 & 0.9773 & 0.9705 & 0.9926 \\
\cmidrule(lr){2-9}
 & \multirow{4}{*}{Mistral 123b} & Abstracts & 0.986 & 0.989 & 0.9895 & 0.984 & 0.9895 & 0.9965 \\
 &  & News & 0.9985 & 0.998 & 0.9995 & 0.9975 & 0.962 & 1.0 \\
 &  & Reviews & 0.9905 & 0.9895 & 0.9945 & 0.993 & 0.9765 & 0.994 \\
 &  & QA & 0.9014 & 0.8333 & 0.9299 & 0.8819 & 0.9167 & 0.9905 \\
\cmidrule(lr){2-9}
 & \multirow{4}{*}{Deepseek 70b} & Abstracts & 0.9885 & 0.989 & 0.992 & 0.9825 & 0.9965 & 0.9885 \\
 &  & News & 0.998 & 0.9985 & 0.998 & 0.9985 & 0.999 & 1.0 \\
 &  & Reviews & 0.9945 & 0.994 & 0.9945 & 0.99 & 0.9955 & 0.9955 \\
 &  & QA & 0.9341 & 0.9019 & 0.9483 & 0.8903 & 0.9151 & 0.954 \\
\midrule
\multirow{28}{*}{RoBERTa} & \multirow{4}{*}{Llama3.3 70b} & Abstracts & 0.9865 & 0.9765 & 0.9925 & 0.9295 & 0.9795 & 0.9835 \\
 &  & News & 0.99 & 0.996 & 0.988 & 0.901 & 0.782 & 0.9965 \\
 &  & Reviews & 0.987 & 0.9925 & 0.9995 & 0.938 & 0.99 & 0.9915 \\
 &  & QA & 0.8481 & 0.8586 & 1.0 & 0.9578 & 0.8291 & 0.9451 \\
\cmidrule(lr){2-9}
 & \multirow{4}{*}{Qwen 14b} & Abstracts & 0.9945 & 0.993 & 0.995 & 0.998 & 0.999 & 0.9965 \\
 &  & News & 0.999 & 1.0 & 0.9995 & 1.0 & 0.9985 & 1.0 \\
 &  & Reviews & 0.9995 & 0.999 & 0.9995 & 1.0 & 0.998 & 1.0 \\
 &  & QA & 0.9826 & 0.9652 & 0.9784 & 0.9852 & 0.9541 & 0.9662 \\
\cmidrule(lr){2-9}
 & \multirow{4}{*}{Qwen 32b} & Abstracts & 0.993 & 0.989 & 0.9935 & 0.993 & 0.999 & 0.9985 \\
 &  & News & 1.0 & 1.0 & 1.0 & 0.999 & 0.9985 & 0.9975 \\
 &  & Reviews & 0.998 & 0.9995 & 0.9965 & 0.9985 & 0.9975 & 0.9955 \\
 &  & QA & 0.9789 & 0.9789 & 0.981 & 0.9873 & 0.9209 & 0.9789 \\
\cmidrule(lr){2-9}
 & \multirow{4}{*}{Qwen 72b} & Abstracts & 0.986 & 0.9825 & 0.9875 & 0.979 & 0.9945 & 0.982 \\
 &  & News & 0.998 & 1.0 & 0.999 & 0.9995 & 0.999 & 0.999 \\
 &  & Reviews & 0.996 & 0.9955 & 1.0 & 0.9945 & 0.998 & 0.9915 \\
 &  & QA & 0.981 & 0.9399 & 0.9831 & 0.9504 & 0.8623 & 0.9699 \\
\cmidrule(lr){2-9}
 & \multirow{4}{*}{Solar 22b} & Abstracts & 0.9825 & 0.987 & 0.9795 & 0.9825 & 0.995 & 0.9995 \\
 &  & News & 0.9975 & 0.995 & 0.994 & 0.9995 & 1.0 & 0.998 \\
 &  & Reviews & 0.9975 & 0.9975 & 1.0 & 0.997 & 0.9935 & 1.0 \\
 &  & QA & 0.9742 & 0.9831 & 0.9963 & 0.9821 & 0.9747 & 0.9937 \\
\cmidrule(lr){2-9}
 & \multirow{4}{*}{Mistral 123b} & Abstracts & 0.9835 & 0.982 & 0.982 & 0.9665 & 0.987 & 0.9925 \\
 &  & News & 0.993 & 0.996 & 0.998 & 0.992 & 0.9825 & 0.999 \\
 &  & Reviews & 0.9955 & 0.981 & 0.9915 & 0.991 & 0.9805 & 0.9935 \\
 &  & QA & 0.8877 & 0.8318 & 0.9383 & 0.9024 & 0.8866 & 0.9826 \\
\cmidrule(lr){2-9}
 & \multirow{4}{*}{Deepseek 70b} & Abstracts & 0.9835 & 0.986 & 0.988 & 0.9655 & 0.996 & 0.9825 \\
 &  & News & 0.997 & 0.986 & 0.9995 & 0.998 & 0.9985 & 0.998 \\
 &  & Reviews & 0.994 & 0.995 & 0.9925 & 0.9925 & 0.9965 & 0.9965 \\
 &  & QA & 0.8813 & 0.8734 & 0.9314 & 0.8645 & 0.8513 & 0.9185 \\
\bottomrule
\end{tabular}}
\caption{Accuracy of the fine-tuned detectors on in-domain data.}\label{tab:results}
\end{table*}

%% file: tables/corr_apen_table.tex
\begin{table*}[htbp]
\centering
\scriptsize
\begin{tabular}{lcccc|ccc}
\toprule
Linsuitic& Feature& \multicolumn{3}{c}{DeBERTa-based}&\multicolumn{3}{c}{RoBERTa-based}\\
 Feature & Metric & C-P &  C-M &  C-D &  C-P &  C-M &  C-D \\
\midrule
 \multirow{3}{*}{Lexical diversity}& MATTR & 0.035 & \textbf{0.312} & 0.097 & 0.031 & \textbf{0.326} & 0.100 \\
 & L MATTR & 0.038 & \textbf{0.313} & 0.087 & 0.021 & \textbf{0.349} & 0.079 \\
  & Unique words & 0.004 & 0.118 & 0.116 & 0.044 & \textbf{0.163} & 0.074 \\
 \midrule
Lexical density & Number of function words & 0.008 & \textbf{0.148} & 0.104 & 0.042 & \textbf{0.282} & 0.184 \\
\midrule
\multirow{6}{*}{Readability} & FLESCH & 0.006 & \textbf{0.254} & \textbf{0.198} & 0.006 & \textbf{0.330} & \textbf{0.251} \\
 & Sentence length & \textbf{0.069} & \textbf{0.242} & 0.093 & \textbf{0.073} & \textbf{0.288} & 0.122 \\
 & Long sentences & 0.020 & \textbf{0.161} & 0.099 & 0.061 & \textbf{0.196} & 0.066 \\
 & Short sentences & \textbf{0.072} & \textbf{0.223} & 0.040 & 0.041 & \textbf{0.290} & 0.043 \\
 & Sentence length std & 0.034 & 0.104 & 0.004 & 0.058 & \textbf{0.210} & 0.064 \\
 & Length in characters & 0.002 & 0.117 & 0.083 & 0.045 & \textbf{0.191} & 0.012 \\
 \midrule
 \multirow{2}{*}{Sentiment} & Polarity & 0.025 & 0.125 & \textbf{0.224} & \textbf{0.106} & \textbf{0.261} & \textbf{0.187} \\
 & Subjectivity & 0.019 & 0.064 & 0.001 & 0.058 & 0.134 & 0.054 \\
\midrule
 & Verbs & 0.050 & 0.080 & 0.005 & \textbf{0.072} & 0.042 & 0.041 \\
 & Nouns & 0.047 & \textbf{0.307} & 0.164 & \textbf{0.066} & \textbf{0.322} & 0.180 \\
 & Adjectives & 0.017 & \textbf{0.147} & 0.038 & 0.037 & 0.014 & 0.092 \\
 & Adverbs & 0.005 & \textbf{0.201} & \textbf{0.217} & 0.016 & \textbf{0.213} & \textbf{0.195} \\
 POS & Determiners & 0.010 & \textbf{0.215} & \textbf{0.218} & 0.044 & \textbf{0.286} & \textbf{0.241} \\
 & Interjections & 0.031 & 0.111 & 0.157 & 0.032 & \textbf{0.169} & 0.096 \\
 & Conjunctions & 0.042 & 0.061 & 0.158 & 0.020 & \textbf{0.190} & 0.098 \\
 & Particles & 0.027 & 0.012 & 0.026 & 0.006 & \textbf{0.212} & 0.047 \\
 & Numerals & 0.008 & \textbf{0.270} & 0.163 & 0.003 & \textbf{0.265} & \textbf{0.202} \\
 & Pronouns & 0.013 & 0.064 & 0.057 & 0.051 & \textbf{0.159} & 0.007 \\
\midrule
 & Content words & 0.023 & 0.128 & 0.064 & 0.039 & \textbf{0.278} & 0.148 \\
 & Function words  & 0.005 & 0.045 & 0.086 & 0.035 & \textbf{0.251} & 0.164 \\
 & Content words types & 0.041 & 0.100 & \textbf{0.195} & 0.001 & \textbf{0.239} & \textbf{0.238} \\
 & Function words types & 0.012 & 0.037 & 0.137 & 0.048 & 0.006 & \textbf{0.194} \\
 & Proper names & \textbf{0.070} & \textbf{0.360} & 0.013 & 0.022 & 0.134 & 0.053 \\
 & Nouns in possessive case& 0.049 & 0.062 & 0.102 & 0.002 & 0.081 & 0.167 \\
 & Adjectives in positive degree & 0.012 & \textbf{0.147} & 0.039 & 0.043 & 0.018 & 0.097 \\
 & Adverbs in positive degree & 0.003 & \textbf{0.188} & \textbf{0.191} & 0.025 & \textbf{0.188} & 0.184 \\
 & Adverbs in comparative degree & 0.006 & \textbf{0.191} & \textbf{0.225} & 0.016 & \textbf{0.206} & \textbf{0.198} \\
 & Adverbs in superlative degree & 0 & \textbf{0.191} & \textbf{0.226} & 0.022 & \textbf{0.204} & \textbf{0.197} \\
 & 'I' pronoun & 0 & \textbf{0.202} & 0.134 & 0.033 & \textbf{0.327} & 0.085 \\
 & 'He' pronoun & 0.045 & \textbf{0.173} & 0.045 & 0.031 & \textbf{0.191} & 0.059 \\
 & 'She' pronoun & \textbf{0.086} & \textbf{0.176} & 0.004 & \textbf{0.080} & \textbf{0.316} & 0.020 \\
 & 'It' pronoun & \textbf{0.067} & 0.092 & 0.167 & \textbf{0.067} & \textbf{0.203} & \textbf{0.199} \\
 & 'You' pronoun & 0.015 & \textbf{0.160} & 0.085 & 0.011 & 0.113 & 0.028 \\
 & 'They' pronoun & 0.015 & 0.072 & 0.101 & 0.018 & 0.128 & 0.128 \\
 & 'Me' pronoun & 0.044 & 0.096 & 0.041 & 0.025 & 0.077 & 0.032 \\
 & 'You' object pronoun & 0.016 & 0.077 & 0.153 & 0.050 & \textbf{0.161} & 0.100 \\
 & 'Him' object pronoun & 0.041 & \textbf{0.161} & 0.069 & 0.042 & 0.034 & 0.097 \\
 Lexical & 'Her' object pronoun & 0.040 & 0.080 & 0.072 & \textbf{0.072} & \textbf{0.203} & 0.031 \\
 & 'Us' pronoun & 0.041 & \textbf{0.215} & 0.101 & 0.012 & 0.086 & 0.137 \\
 & 'Them' pronoun & 0.052 & 0.030 & 0.073 & \textbf{0.080} & 0.080 & 0.012 \\
 & 'My' pronoun & 0.021 & \textbf{0.214} & 0.127 & 0.053 & \textbf{0.341} & 0.109 \\
 & 'Your' pronoun & 0.007 & \textbf{0.250} & 0.085 & 0.009 & \textbf{0.254} & 0.096 \\
 & 'His' pronoun & 0.005 & 0.078 & 0.086 & 0.023 & 0.059 & 0.152 \\
 & 'Her' possessive pronoun & 0.034 & 0.004 & 0.053 & \textbf{0.063} & 0.102 & 0.129 \\
 & 'Its' possessive pronoun & \textbf{0.089} & \textbf{0.293} & \textbf{0.245} & \textbf{0.079} & \textbf{0.323} & \textbf{0.244} \\
 & 'Their' possessive pronoun & \textbf{0.065} & \textbf{0.237} & 0.121 & 0.015 & \textbf{0.228} & 0.132 \\
 & 'Yours' pronoun& 0.026 & 0.001 & \textbf{0.221} & 0.009 & 0.103 & 0.163 \\
 & 'Theirs' pronoun & 0.004 & 0.011 & \textbf{0.255} & 0.008 & 0.084 & \textbf{0.237} \\
 & 'Hers' pronoun & 0.061 & 0.006 & 0.067 & 0.031 & 0.130 & 0.021 \\
 & 'Ours' possessive pronoun & 0.003 & 0.106 & 0.021 & 0.010 & \textbf{0.186} & 0.005 \\
 & 'Myself' pronoun & 0.023 & \textbf{0.157} & 0.088 & 0.017 & \textbf{0.297} & 0.073 \\
 & 'Himself' pronoun & 0.012 & \textbf{0.332} & 0.052 & 0.027 & \textbf{0.232} & 0.043 \\
 & 'Herself' pronoun & 0.036 & \textbf{0.248} & 0.025 & 0.032 & \textbf{0.227} & 0.001 \\
 & 'Itself' pronoun & 0.032 & 0.132 & \textbf{0.221} & 0.055 & 0.027 & \textbf{0.204} \\
 & 'Ourselves' pronoun & 0 & 0.103 & 0.073 & 0.018 & \textbf{0.269} & 0.068 \\
 & 'Yourselves' pronoun & 0.018 & \textbf{0.177} & 0.108 & 0.005 & \textbf{0.194} & 0.083 \\
 & 'Themselves' pronoun & \textbf{0.064} & \textbf{0.230} & 0.019 & \textbf{0.100} & \textbf{0.278} & 0.015 \\
 & First person singular pronouns & 0 & \textbf{0.202} & 0.134 & 0.033 & \textbf{0.327} & 0.085 \\
 & Second person pronouns & 0.014 & \textbf{0.204} & 0.056 & 0.011 & \textbf{0.192} & 0.005 \\
 & Third person singular pronouns & 0.047 & 0.014 & 0.093 & 0.031 & \textbf{0.198} & 0.140 \\
 & Third person plural pronouns & 0.050 & \textbf{0.182} & 0.117 & 0.039 & 0.129 & 0.157 \\
 \midrule
 	General & Incidence of verbs in infinitive & 0.037 & 0.050 & 0.020 & \textbf{0.095} & \textbf{0.153} & 0.085 \\
\bottomrule
\end{tabular}
\caption{The correlation between generalization performance with different linguistic features. The \textbf{significant correlation is bolded}. }
\label{tab:app-corr}
\end{table*}